\documentclass[onefignum,onetabnum]{siamart220329}
\usepackage{amsmath,amssymb}
\usepackage{graphicx}
\usepackage{float}
\usepackage{array}
\allowdisplaybreaks
\floatstyle{ruled}\newfloat{algorithm}{htbp}{loa}\floatname{algorithm}{Algorithm}\floatstyle{plain}
\newcommand{\Pset}{\mathcal{P}}          
\newcommand{\Lam}{\Lambda}               
\newcommand{\kmin}{\kappa_{\min}}
\newcommand{\diag}{\operatorname{diag}}
\newcommand{\T}{^{\mathsf T}}

\usepackage{cleveref}
\hypersetup{hypertexnames=false}
\usepackage{booktabs}
\usepackage{placeins}
\crefname{figure}{Figure}{Figures}\crefname{table}{Table}{Tables}
\usepackage{etoolbox}
\AtBeginEnvironment{lemma}{\crefalias{theorem}{lemma}}
\AtBeginEnvironment{proposition}{\crefalias{theorem}{proposition}}
\AtBeginEnvironment{corollary}{\crefalias{theorem}{corollary}}
\AtBeginEnvironment{remark}{\crefalias{theorem}{remark}}
\AtBeginEnvironment{example}{\crefalias{theorem}{example}}
\AtBeginEnvironment{definition}{\crefalias{theorem}{definition}}
\crefname{definition}{Definition}{Definitions}\crefname{algorithm}{Algorithm}{Algorithms}\Crefname{algorithm}{Algorithm}{Algorithms}\Crefname{definition}{Definition}{Definitions}
\newsiamremark{remark}{Remark}
\newsiamremark{example}{Example}
\crefname{lemma}{Lemma}{Lemmas}\Crefname{lemma}{Lemma}{Lemmas}
\crefname{proposition}{Proposition}{Propositions}\Crefname{proposition}{Proposition}{Propositions}
\crefname{corollary}{Corollary}{Corollaries}\Crefname{corollary}{Corollary}{Corollaries}
\crefname{theorem}{Theorem}{Theorems}\Crefname{theorem}{Theorem}{Theorems}
\crefname{remark}{Remark}{Remarks}\crefname{example}{Example}{Examples}

\title{Which Ellipses Are Outlines of a Given Ellipsoid?\\ A Ratio Test with Unknown Focal Length}
\author{Michael Werman\thanks{School of Computer Science and Engineering, The Hebrew University of Jerusalem (\email{michael.werman@mail.huji.ac.il}).
}}
\headers{Which Ellipses Are Outlines of a Given Ellipsoid?}{M. Werman}

\begin{document}
\maketitle

\begin{abstract}
When is an image ellipse the outline of a given ellipsoid with semi-axes $a\ge b\ge c$? For a camera with square pixels, zero skew and known principal point, we show that it is an outline at focal length $f$ exactly when a cone ratio $\kappa(f)$, computed from two eigenvalues, is at most $a/c$; the middle semi-axis $b$ plays no role. With $f$ unknown, the admissible focal lengths form an interval; whether it is empty is decided by the smallest cone ratio over $f$, which, when attained, is obtained from an explicit quartic. The interval is bounded above only when the principal point is closer to the long axis of the ellipse than to its short axis. Thus one ellipse can rule out ellipsoids that are too round, never ones that are too elongated, and none if the principal point is unknown. Simulations and 45 Cassini images of Saturn's moons support the test.
\end{abstract}

\begin{keywords}
ellipsoid outlines, apparent contour, perspective projection, focal length, confocal quadrics, tangent cones, camera calibration
\end{keywords}

\begin{MSCcodes}
68T45, 68U10, 51N05, 51N15, 15A18, 15A42
\end{MSCcodes}

\section{Introduction}\label{sec:intro}
Ellipsoids model objects that appear as image ellipses in object-level mapping \cite{Rubino2018,Nicholson2019}, relocalization \cite{GaudilliereP2E2020,Gaudilliere2023,ZinsSimonBerger2022}, and horizon-based navigation and calibration \cite{Christian2021,DanasRiveraPeck2023}. Before computing a pose, one may ask three questions. Can this ellipse be the outline of this ellipsoid? If so, at which focal lengths? And what does a single ellipse reveal about the shape? The answers can be used to reject wrong hypotheses or correspondences early, to bound an unknown focal length (zoom lenses, archival images), and to initialize pose solvers. These are questions of realizability, not reconstruction: we ask whether some pose and focal length produce the ellipse, not which one did.

\paragraph{The ratio test} Let the ellipsoid have semi-axes $a\ge b\ge c$, and let the camera have square pixels, zero skew, no lens distortion and known principal point $p$. In homogeneous pixel coordinates $\mathbf{x}=(u,v,1)\T$ centered at $p$, write the image ellipse $E$ as $\mathbf{x}\T C\mathbf{x}=0$, with $C$ signed so that its upper-left $2\times2$ block is positive definite. For a focal length $f$, put $K_f=\diag(f,f,1)$ and $Q_f=K_fCK_f$. The rays from the camera center through the points of $E$ form the \emph{cone of rays} $d\T Q_fd=0$, where $d=(u/f,v/f,1)\T$ is the direction of the ray through $(u,v)$. The matrix $Q_f$ has two positive eigenvalues $\mu_-\le\mu_+$, which depend on the scale of $C$; their \emph{cone ratio} $\kappa(f)=\sqrt{\mu_+/\mu_-}$ does not. Our first result (\cref{thm:ellipsoid}) is
\[
E \text{ is an outline at focal length } f \iff \kappa(f)\le a/c .
\]

\paragraph{Assumptions} The theory needs a central projection and the exact conic of the outline, although the outline itself need not be fully visible. Any known calibration reduces to square pixels and zero skew (\cref{rem:generalK}). In practice the conic is fitted to noisy edges, predicted by a detector \cite{ZinsSimonBerger2022} or, for moons, fitted to the sunlit limb; \cref{rem:certified} handles a bounded error in it.

\paragraph{Contributions}
\begin{enumerate}
\item \emph{The ratio test} (\cref{sec:test}): one symmetric $3\times3$ eigenvalue problem, independent of $b$; its constructive proof gives the poses (\cref{rem:pose}).
\item \emph{Where the camera can be} (\cref{sec:where}): one inequality between two confocal coordinates of the camera center with respect to $E$.
\item \emph{Unknown focal length} (\cref{sec:focal}), our main contribution: $\kappa(f)$ is quasiconvex, so the admissible $f$ form an interval (\cref{thm:shape}); its emptiness is decided by $\kmin$, obtained when attained from a quartic (\cref{thm:kminform}); it is bounded above only in the long-axis wedge (\cref{prop:wedge}); \cref{alg:test} computes it.
\item \emph{What one ellipse reveals} (\cref{sec:practical}): only too-round ellipsoids are ruled out, none if $p$ is unknown; error bounds for noisy ellipses; general calibrations.
\item \emph{Experiments} (\cref{sec:numerics}): simulations and 45 Cassini images.
\end{enumerate}
Proofs are in \cref{app:ellipsoids}.

\paragraph{Related work} Pose from conics and quadric outlines goes back to \cite{FerriMangiliViano1993,Ma1993}; pose from one ellipse has been computed for spheroids \cite{WokesPalmer2010} and ellipsoids \cite{Gaudilliere2023}, whose existence condition \cite[Thm.~1]{Gaudilliere2023}, with a free scalar parameter, is equivalent to our test for a calibrated camera (\cref{app:gsb}); \cref{prop:folds} corrects a remark there on circular tangent cones.

Sphere outlines are well studied \cite{Daucher1994,AgrawalDavis2003,YingZha2006,ZhangWongZhang2007}; with $p$ known, one sphere image fixes $f$ \cite{Penne2015,HanWongTan2016,OzgurAlkhatibMezouarBartoli2026}, which is the case $a=c$ of our test. With the pose known, one image of an ellipsoid can fix $f$ and $p$ \cite{DanasRiveraPeck2023}; we assume the pose unknown and $p$ known. One outline of a generic algebraic surface of degree at least three determines it up to projectivity \cite{Forsyth1996}; quadrics are the exception.

The geometry is classical: tangent cones of quadrics are described by confocal quadrics \cite{Salmon1882}, \cite[Ch.~VIII]{Smith1907}, the cone over an ellipse is circular exactly from its focal hyperbola \cite{GlaeserStachelOdehnal2016}, and the tangent cone from $O$ has dual matrix $D-OO\T$, $D=\diag(a^2,b^2,c^2)$ \cite[Sec.~8.3]{HartleyZisserman2004}, so interlacing \cite{Golub1973} gives $\kappa\le a/c$, in the orthographic limit the classical range of projected axis ratios \cite{Stark1977,Binney1985}. New here are the converse of this bound, in closed form and without a free parameter, the region of camera centers, and the analysis with $f$ unknown.

\section{Setting and notation}\label{sec:setting}
\begin{figure}[htbp]
\centering\includegraphics[width=0.42\linewidth]{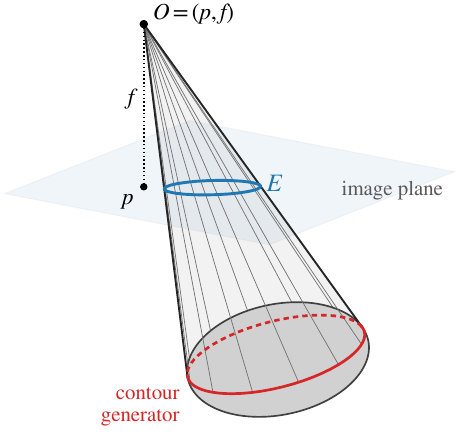}
\caption{The rays from the camera center $O$ that are tangent to the ellipsoid form its tangent cone (shaded). They touch the ellipsoid along the contour generator (red) and meet the image plane in the outline $E$ (blue). The perspective scene is drawn in an orthographic view.}\label{fig:setting}
\end{figure}

The camera is as in \cref{sec:intro}; its focal length $f$ is given in \cref{sec:test} and unknown from \cref{sec:focal} on. With the image plane at $z=0$, the camera center is $O=(p,f)$, at height $f$ above $p$ (\cref{fig:setting}). The ellipsoid has unknown pose and known shape: its semi-axes $a\ge b\ge c$ are known up to a common factor, since one image cannot separate size from distance. It is \emph{triaxial} if $a>b>c$, and a \emph{spheroid} if two semi-axes are equal: \emph{prolate} if $a>b=c$, \emph{oblate} if $a=b>c$.

The image ellipse $E$ has semi-axes $a_E\ge b_E$ and conic matrix
\[
C=\begin{pmatrix}A_E&\beta_E\\ \beta_E\T&\gamma_E\end{pmatrix}
\]
in coordinates centered at $p$, signed so that $A_E$ is positive definite; its eigenvalues $0<\alpha_1\le\alpha_2$ satisfy $\alpha_2/\alpha_1=a_E^2/b_E^2$. The entry $\gamma_E$ is positive, zero or negative as $p$ lies outside, on or inside $E$. Since $E$ is a real nondegenerate ellipse, $\det C=\det A_E\,(\gamma_E-\beta_E\T A_E^{-1}\beta_E)<0$, so $C$ has one negative and two positive eigenvalues, and, by Sylvester's law of inertia, so does $Q_f$ for every $f>0$. \Cref{tab:notation} lists the main notation.

\begin{table}[tbp]
\centering\small
\caption{Main notation; other symbols are defined where they are used.}\label{tab:notation}
\vspace{3pt}
\begin{tabular}{@{}>{\raggedright\arraybackslash}p{0.27\linewidth}>{\raggedright\arraybackslash}p{0.67\linewidth}@{}}
\toprule
$a\ge b\ge c$; $\Lam$ & semi-axes of the ellipsoid; $\diag(a^{-2},b^{-2},c^{-2})$\\
$E$, $a_E\ge b_E$; $C$ & image ellipse and its semi-axes; its conic matrix, centered at $p$, with blocks $A_E$ (eigenvalues $\alpha_1\le\alpha_2$), $\beta_E$, $\gamma_E$\\
$p$, $f$, $O$; $K_f$, $Q_f$ & principal point, focal length, camera center; $\diag(f,f,1)$, $K_fCK_f$ (cone of rays)\\
$(x,y)$; $d_E$, $e_E$ & position of $p$ in the frame of $E$; distances from the center of $E$ to $p$ and to a focus\\
$-\mu_0<0<\mu_-\le\mu_+$; $\kappa$ & eigenvalues of $Q_f$; cone ratio $\sqrt{\mu_+/\mu_-}$\\
$\kmin$, $f^*$; $\kappa_\infty$, $\kappa_0$ & $\inf_f\kappa$ and, when the infimum is attained, its minimizer; limits of $\kappa$ as $f\to\infty$ ($=a_E/b_E$) and $f\to0$ ($p$ inside $E$)\\
$\lambda_i$; $\theta_i$ & confocal coordinates of $O$ with respect to the ellipsoid; to $E$\\
$\rho$; $I(\rho)$ & hypothesized $a/c$; interval $\{f>0:\kappa(f)\le\rho\}$ of admissible focal lengths\\
\bottomrule
\end{tabular}
\end{table}

\begin{definition}[outline]\label{def:outline}
$E$ is an \emph{outline} (or apparent contour) of the ellipsoid at focal length $f$ if the ellipsoid can be rotated, translated and uniformly scaled into the open half-space $z<f$, in front of the camera's \emph{principal plane} $z=f$, so that the boundary of its central projection from $O$ onto the image plane $z=0$ is $E$.
\end{definition}

\paragraph{The frame of $E$} We place the origin at the center of $E$, the first axis along its long axis and the third axis vertical, so that $O=(x,y,f)$, where $(x,y)$ is the position of $p$. The \emph{axis lines} of $E$ are its two lines of symmetry. The \emph{long-axis wedge} of $E$ is the region $|x|>|y|$: the points closer to the long axis line than to the short axis line. Let $d_E$ be the distance from the center of $E$ to $p$ and, for $a_E>b_E$, let $e_E=\sqrt{a_E^2-b_E^2}$ be the distance from the center to a focus. The \emph{focal hyperbola} of $E$ is $x^2/e_E^2-z^2/b_E^2=1$, $y=0$; it passes through the foci of $E$, and the cone of rays through $E$ is circular ($\kappa=1$) exactly when the camera center lies on it \cite{GlaeserStachelOdehnal2016} (see also \cref{prop:confocal,lem:sphere}).

\section{The ratio test}\label{sec:test}
Let $-\mu_0<0<\mu_-\le\mu_+$ be the eigenvalues of $Q_f$, the matrix of the cone of rays from the camera center through $E$. Its \emph{cone axis} is the eigenvector of $-\mu_0$, and its \emph{cone ratio} $\kappa=\sqrt{\mu_+/\mu_-}\ge1$ is the ratio of the semi-axes of its cross-sections perpendicular to the cone axis; in general $\kappa\ne a_E/b_E$. In this section we work in the ellipsoid frame and also write $O$ for the camera center in that frame.

\subsection{Tangent cones}
Through each point outside the ellipsoid pass three quadrics confocal with it; their parameters are the \emph{confocal coordinates} of the point. Classically, the tangent cone from the point has its axes along the normals of these quadrics, and its eigenvalues are determined by the confocal coordinates \cite{Salmon1882}, \cite[Ch.~VIII]{Smith1907}. This reduces the test to bounds on confocal coordinates.

\begin{lemma}[tangent cones and confocal quadrics]\label{lem:cone}
Let the ellipsoid be $X\T\Lam X=1$ with $a\ge b\ge c$, and let $O=(O_1,O_2,O_3)$ lie outside it, so that $\varsigma=O\T\Lam O-1>0$. Then:
\begin{enumerate}
\item[(i)] the tangent cone from $O$ is $d\T Md=0$, with $M=\varsigma\Lam-\Lam OO\T\Lam$;
\item[(ii)] the equation
\begin{equation}\label{eq:confocal}
\frac{O_1^2}{a^2+\lambda}+\frac{O_2^2}{b^2+\lambda}+\frac{O_3^2}{c^2+\lambda}=1,
\end{equation}
with denominators cleared, has three real roots, the confocal coordinates $\lambda_1>0$, $-b^2\le\lambda_2\le-c^2$ and $-a^2\le\lambda_3\le-b^2$ of $O$, with strict inequalities when $a>b>c$ and $O$ lies off the coordinate planes;
\item[(iii)] $M$ has eigenvalues $-\varsigma/\lambda_i$, $i=1,2,3$. Hence its cone ratio is $\sqrt{\lambda_3/\lambda_2}\in[1,a/c]$.
\end{enumerate}
\end{lemma}

When the ellipsoid lies in front of the principal plane (\cref{def:outline}), the image plane cuts its tangent cone from $O$ in an ellipse, the outline $E$. The cone touches the ellipsoid along the contour generator (\cref{fig:setting}), the ellipse in which the polar plane of $O$ meets the ellipsoid \cite[Sec.~8.3]{HartleyZisserman2004}.

\subsection{The test}
By \cref{lem:cone}, no tangent cone has cone ratio above $a/c$. The converse, proved in \cref{app:ellipsoids}, is that every real elliptic cone with cone ratio at most $a/c$ is, after a rigid motion and up to a positive factor, a tangent cone of the ellipsoid. Together they give the test.

\begin{theorem}[the ratio test]\label{thm:ellipsoid}
Let the ellipsoid have semi-axes $a\ge b\ge c$, and let the camera have square pixels, zero skew and known principal point $p$. Let $C$ be the conic matrix of $E$ in coordinates centered at $p$, signed as in \cref{sec:setting}, so that $Q_f=K_fCK_f$, $K_f=\diag(f,f,1)$, has two positive eigenvalues $\mu_-\le\mu_+$, and let $\kappa(f)=\sqrt{\mu_+/\mu_-}$. Then:
\begin{enumerate}
\item[(i)] $E$ is an outline of the ellipsoid at focal length $f$ (\cref{def:outline}) if and only if $\kappa(f)\le a/c$;
\item[(ii)] for fixed $p$ and $f$, the set of outlines depends on $(a,b,c)$ only through $a/c$.
\end{enumerate}
\end{theorem}

The converse is the substantial direction of (i): every real elliptic cone satisfying the bound is a tangent cone of the given ellipsoid, and the proof constructs the pose (\cref{rem:pose}). By (ii), for example, a prolate and an oblate spheroid with the same $a/c$, such as a rugby ball and a lentil, have the same outlines.

\paragraph{Why $b$ drops out} Up to rotation and scale, an elliptic cone has two invariants: its squared cone ratio $\mu_+/\mu_-$ and its \emph{aperture} $\mu_0/\mu_-$. For a tangent cone these are $\lambda_3/\lambda_2$ and $|\lambda_3|/\lambda_1$ (\cref{lem:cone}). Since $|\lambda_2|$ ranges over $[c^2,b^2]$ and $|\lambda_3|$ over $[b^2,a^2]$, the ratio $\lambda_3/\lambda_2$ takes every value in $[1,a^2/c^2]$, whatever $b$ is; the free coordinate $\lambda_1>0$ then matches the aperture, and the proof of \cref{thm:ellipsoid} gives the point $O$ with these coordinates in closed form. Thus $b$ drops out of the set of outlines, but not out of the poses that produce them (\cref{rem:pose}).

\begin{remark}[special cases]\label{rem:special}
\emph{Spheres.}\label{lem:sphere} If $a=c$, the test reads $\kappa=1$. With \cref{prop:confocal} below this gives the classical description of sphere outlines \cite{Daucher1994,Penne2015,HanWongTan2016}: $E$ is an outline of a sphere at focal length $f>0$ if and only if either $a_E>b_E$, $p$ lies on the long axis line of $E$ beyond a focus ($d_E>e_E$) and $f^2=b_E^2(d_E^2-e_E^2)/e_E^2$, that is, the camera center lies on the focal hyperbola of $E$; or $E$ is a circle centered at $p$, and $f>0$ is arbitrary. Hence, if $p$ may be chosen, every ellipse is a sphere outline at every $f>0$. Since $a/c\ge1$, a sphere outline is, for the same $p$ and $f$, an outline of every ellipsoid.

\emph{Orthographic limit.} As $f\to\infty$, $\kappa\to a_E/b_E$, and the test becomes $a_E/b_E\le a/c$, the classical range of axis ratios of orthographic outlines \cite{Stark1977,Binney1985}; the supplementary materials derive it directly by interlacing.
\end{remark}

\begin{remark}[recovering the pose]\label{rem:pose}
The proof of \cref{thm:ellipsoid} is constructive. Let $a>b>c$ and $\kappa(f)\le a/c$ for a known $f$. The interval $[\max(c^2\mu_+,b^2\mu_-),\min(b^2\mu_+,a^2\mu_-)]$ is then nonempty. For each $\tau$ in it, Step~2 of that proof, applied to $(\lambda_1,\lambda_2,\lambda_3)=(\tau/\mu_0,-\tau/\mu_+,-\tau/\mu_-)$, gives a camera center in the ellipsoid frame, and Step~3 gives the rotation. The poses form a one-parameter family when $1<\kappa<a/c$; when $\kappa=a/c$ the pose is determined up to the symmetries of the ellipsoid, and when $\kappa=1$ up to a rotation about the cone axis.
\end{remark}

\begin{figure}[htbp]
\centering\includegraphics[trim=0 0 325.5 0,clip,width=0.4\linewidth]{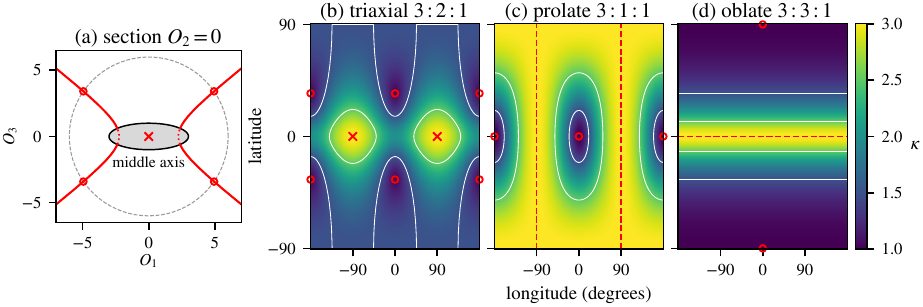}
\caption{The $3{:}2{:}1$ ellipsoid cut by the plane $O_2=0$. From the points of its focal hyperbola (red; dotted inside the ellipsoid) the tangent cone is circular; $\circ$: where the hyperbola meets a sphere of radius $6$ (dashed). $\times$: the middle axis (normal to this section), from whose points outside the ellipsoid $\kappa=a/c$. Maps of $\kappa$ over viewing spheres are in the supplementary materials.}\label{fig:directions}
\end{figure}

\subsection{Extreme viewpoints}
For fixed $f$, the smooth \emph{image map} sends a placement of the ellipsoid in front of the principal plane (center, rotation and uniform scale: a $7$-dimensional manifold) to the conic of its outline, a point of the $5$-dimensional manifold of real ellipses (conic matrices up to a positive factor).

\begin{proposition}[where $\kappa=a/c$ and $\kappa=1$]\label{prop:folds}
For a triaxial ellipsoid, $\kappa=a/c$ exactly from the points of the middle axis outside it ($O_1=O_3=0$, $|O_2|>b$), and for every fixed $f$ the outlines seen from these points are critical values of the image map; $\kappa=1$ exactly from those points outside it that lie on its \emph{focal hyperbola} $O_2=0$, $O_1^2/(a^2-b^2)-O_3^2/(b^2-c^2)=1$ \cite[Ch.~VIII]{Smith1907}. For a spheroid, $\kappa=a/c$ exactly from the points of its equatorial plane outside it, and $\kappa=1$ exactly from the points of its symmetry axis outside it.
\end{proposition}

In particular, a triaxial ellipsoid has circular tangent cones (\cref{fig:directions}), contrary to a remark in \cite[Sec.~6.1]{Gaudilliere2023}; the supplementary materials give an explicit example and locate the gap in that remark.

\section{Where the camera can be}\label{sec:where}
The test depends on $p$ and $f$ only through the camera center $O=(p,f)$, so it defines a region of admissible camera centers, which we now describe. In the frame of $E$, let $E$ be $x^2/a_E^2+y^2/b_E^2=1$ in the plane $z=0$, with $a_E>b_E$. Through each point $O$ above the image plane pass three quadrics of the confocal family
\[
\frac{x^2}{a_E^2+\theta}+\frac{y^2}{b_E^2+\theta}+\frac{z^2}{\theta}=1,
\]
whose limit as $\theta\to0^+$ is $E$: an ellipsoid ($\theta_1>0$), a hyperboloid of one sheet ($-b_E^2\le\theta_2<0$) and a hyperboloid of two sheets ($-a_E^2\le\theta_3\le-b_E^2$). We call $\theta_1,\theta_2,\theta_3$ the confocal coordinates of $O$ with respect to $E$. On the planes $y=0$ and $x=0$, one of the three quadrics degenerates to that plane.

\begin{proposition}[the camera center in confocal coordinates]\label{prop:confocal}
Let $a_E>b_E$, and let $O=(p,f)$ have confocal coordinates $\theta_1,\theta_2,\theta_3$ with respect to $E$. Then $Q_f$ has $\kappa^2=\theta_3/\theta_2$, so $E$ is an outline at focal length $f$ of an ellipsoid with $a/c=\rho$ exactly when $|\theta_3|\le\rho^2|\theta_2|$. The set of such $O$ lies between two confocal hyperboloids of one sheet, the inner one $\theta_2=-a_E^2/\rho^2$ and the outer one $\theta_2=-b_E^2/\rho^2$:
\[
\{\theta_2\le-a_E^2/\rho^2\}\subseteq\{|\theta_3|\le\rho^2|\theta_2|\}\subseteq\{\theta_2\le-b_E^2/\rho^2\}
\]
(the first set is empty when $\rho<a_E/b_E$). Its boundary $\theta_3=\rho^2\theta_2$ meets the inner hyperboloid in the plane $x=0$ and the outer one in the plane $y=0$; for $\rho>1$ it is not itself a confocal quadric. Further, $\kappa=1$ exactly on the focal hyperbola of $E$, where $\theta_2=\theta_3=-b_E^2$; and as $f\to\infty$, $\theta_2\to-b_E^2$ and $\theta_3\to-a_E^2$, so $\kappa\to a_E/b_E$.
\end{proposition}

For a circle $E$, which \cref{prop:confocal} excludes, $\kappa=1$ exactly when $O$ lies above its center.

\begin{figure}[htbp]
\centering\includegraphics[width=\linewidth]{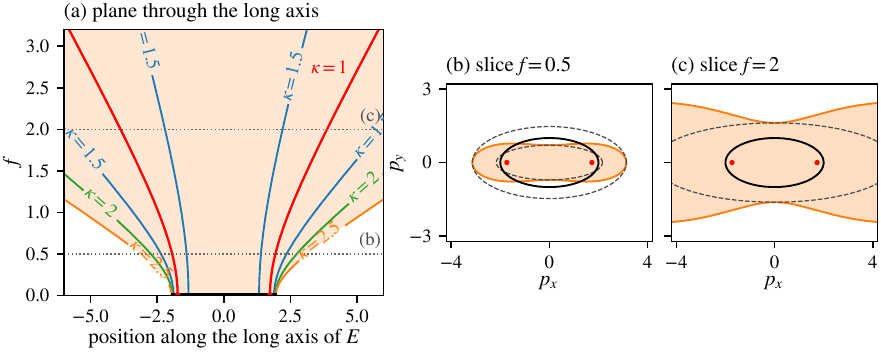}
\caption{Where the camera can be, for $E$ with semi-axes $2$ and $1$ (black). (a) The vertical plane through the long axis of $E$: level curves of $\kappa$, the focal hyperbola ($\kappa=1$, red) and the region $\kappa\le2.5$ (shaded). (b), (c) The horizontal slices $z=f=0.5$ and $z=f=2$ marked in (a). The admissible principal points for an ellipsoid with $a/c=2.5$ (shaded) lie between the two hyperboloids of \cref{prop:confocal} (dashed). In (c), the slice of the outer one, an ellipse with semi-axes $9.99$ and $4.67$, lies outside the frame. Red dots: the foci of $E$.}\label{fig:confocal}
\end{figure}

\Cref{fig:confocal}(a) shows the vertical plane through the long axis of $E$. There one negative confocal coordinate is $-b_E^2$, and the other, $\theta\in[-a_E^2,0)$, labels the hyperbola $x^2/(a_E^2+\theta)-z^2/(-\theta)=1$, whose foci are the ends of the long axis. In this plane $\kappa^2=\max(-b_E^2/\theta,-\theta/b_E^2)$, so $\kappa\le\rho$ exactly when $\max(-a_E^2,-\rho^2b_E^2)\le\theta\le-b_E^2/\rho^2$.

\section{Unknown focal length}\label{sec:focal}
With $f$ unknown, the question becomes which focal lengths pass the test of \cref{thm:ellipsoid}. For $\rho\ge1$ let
\[
I(\rho)=\{f>0:\kappa(f)\le\rho\},
\]
the interval (by \cref{thm:shape}) of admissible focal lengths for an ellipsoid with $a/c=\rho$. We study the function $\kappa(f)$: its limits and infimum (\cref{prop:kmin}), its shape (\cref{thm:shape}), its behavior for large $f$ (\cref{prop:wedge}), and its minimum (\cref{sec:choosef}); \cref{alg:test} combines them. Many statements need $E$ not to be a circle and $p$ to lie on neither axis line of $E$; we then say that $p$ is \emph{off-axis}.

\subsection{The smallest cone ratio and the shape of \texorpdfstring{$\kappa(f)$}{kappa(f)}}

\begin{proposition}[how elongated the ellipsoid must be]\label{prop:kmin}
Let $\kmin(E,p)=\inf_{f>0}\kappa(f)$, written $\kmin$ when $E$ and $p$ are fixed. Some focal length makes $E$ an outline of an ellipsoid with semi-axes $a\ge b\ge c$ if $a/c>\kmin$, none does if $a/c<\kmin$, and, if $a/c=\kmin$, one does exactly when the infimum is attained. Moreover:
\begin{enumerate}
\item[(i)] As $f\to\infty$, $\kappa\to\kappa_\infty=a_E/b_E$. As $f\to0$, $\kappa\to\infty$ if $p$ is outside or on $E$; if $p$ is inside $E$ ($\gamma_E<0$), then $\kappa\to\kappa_0=\sqrt{s_+/s_-}$, where $s_+\ge s_-$ are the eigenvalues of the positive definite matrix $S=A_E-\beta_E\beta_E\T/\gamma_E$. We call $\kappa_\infty$ and $\kappa_0$ the \emph{end limits} of $\kappa$.
\item[(ii)] $\kappa(f)=1$ for some $f$ if and only if $E$ is a sphere outline (\cref{lem:sphere}). The infimum is $\kmin=1$ but is not attained exactly when $E$ is a circle not centered at $p$ (then $\kappa\to1$ as $f\to\infty$) or $p$ is a focus of a noncircular $E$ (then $\kappa\to1$ as $f\to0$).
\end{enumerate}
\end{proposition}

\begin{theorem}[the shape of $\kappa(f)$]\label{thm:shape}
Let $p$ not be the center of $E$ (where $\kappa\equiv\kappa_\infty$).
\begin{enumerate}
\item[(i)] \emph{Interval.} For every $\rho\ge1$, $I(\rho)$ is an interval, possibly empty; equivalently, $\kappa$ is quasiconvex in $f$.
\item[(ii)] \emph{Shape.} $\kappa$ has at most one critical point. If it has one, $f^*$, then $\kappa$ decreases strictly on $(0,f^*]$ and increases strictly on $[f^*,\infty)$, and $\kappa(f^*)$ is below $\kappa_\infty$ and, if $p$ is inside $E$, below $\kappa_0$. Otherwise $\kappa$ is strictly monotone, except for a sphere outline, where it has a corner minimum $\kappa=1$. If $p$ lies on an axis line of $E$, or $E$ is a circle, $\kappa$ has no critical point.
\item[(iii)] \emph{Attainment.} Let $p$ be off-axis. Then $\kmin$ is attained if and only if $p$ lies in the long-axis wedge ($|x|>|y|$) and, if $p$ is inside $E$, also $\beta_+^2>\beta_-^2$, where $\beta_\pm$ are the components of $\beta_E$ in an orthonormal eigenbasis of $S$ for $s_\pm$. If $\kmin$ is not attained, $\kappa$ is strictly monotone and $\kmin$ is the smaller end limit.
\end{enumerate}
\end{theorem}

\begin{figure}[htbp]
\centering\includegraphics[width=\linewidth]{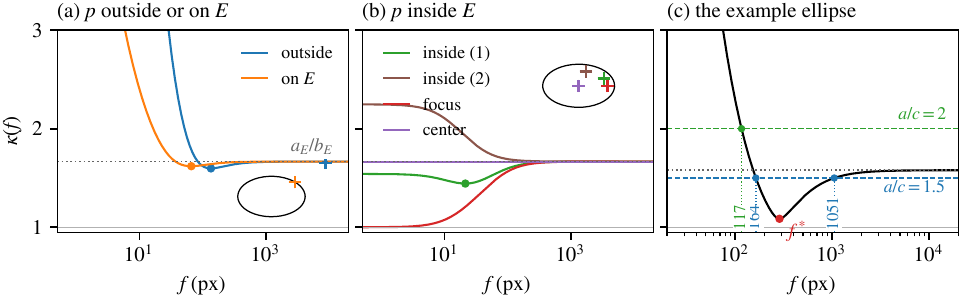}
\caption{(a), (b) $\kappa(f)$ for an ellipse with semi-axes $50$ and $30$~px and six principal points (marked $+$ in the insets). Each curve tends to $a_E/b_E$ (dotted) as $f\to\infty$ and has at most one interior minimum (dots). (a) $p$ outside or on $E$: $\kappa\to\infty$ as $f\to0$. (b) $p$ inside $E$: $\kappa$ tends to a finite $\kappa_0$ as $f\to0$; when $p$ is a focus, $\kappa_0=1$ but $\kappa>1$ for every $f$. (c) The ellipse of \cref{ex:ellipse}: $\kmin\approx1.0863$ at $f^*\approx287.2$~px; $I(1.5)=[164.1,1051.4]$~px and $I(2)=[116.8,\infty)$~px.}\label{fig:gallery}
\end{figure}

By (i) and the end limits, $I(\rho)$ is bounded above if $\rho<\kappa_\infty$ and unbounded above if $\rho>\kappa_\infty$; it is bounded away from $0$ if $p$ is outside or on $E$, and has lower end $0$ if $p$ is inside $E$ and $\rho>\kappa_0$ (\cref{fig:gallery}). The conditions in (iii) say that, near each end, $\kappa$ falls below its end limit. For $f\to\infty$ this is the long-axis-wedge condition, by the expansion in \cref{prop:wedge}; for $f\to0$ and $p$ inside $E$ it is $\beta_+^2>\beta_-^2$, by the expansion
\[
\kappa(f)^2=\kappa_0^2\Bigl(1-\frac{\beta_+^2-\beta_-^2}{\gamma_E^2}f^2\Bigr)+O(f^4),
\]
which follows by first-order perturbation: as $f$ grows from $0$, the positive eigenvalues of $Q_f/f^2$ decrease from $s_\pm$ by amounts proportional to $s_\pm\beta_\pm^2$, so $\kappa$ first decreases exactly when $\beta_E$ has the larger component along the eigenvector for $s_+$. If $\beta_+^2=\beta_-^2$, the $f^2$ term vanishes, and $\kmin$ is again not attained (\cref{app:ellipsoids}).

\subsection{When the focal length is bounded above}

\begin{proposition}[when $f$ is bounded above]\label{prop:wedge}
Let $E$ not be a circle, let $p$ not be its center, and let $(x,y)$ be the position of $p$ in the frame of $E$ (\cref{sec:setting}). As $f\to\infty$,
\[
\kappa(f)^2=\kappa_\infty^2\Bigl(1+\frac{y^2-x^2}{f^2}\Bigr)+O(f^{-4}).
\]
\begin{enumerate}
\item[(i)] If $p$ lies in the long-axis wedge of $E$ ($|x|>|y|$), then $\kmin<\kappa_\infty$, and $I(\rho)$ is nonempty and bounded above for every $\rho\in(\kmin,\kappa_\infty)$.
\item[(ii)] If $|x|\le|y|$ (which includes the short axis line $x=0$), then $\kappa(f)>\kappa_\infty$ for every $f$. Hence $I(\rho)$ is empty for $\rho\le\kappa_\infty$ and unbounded above for $\rho>\kappa_\infty$, and with $f$ unknown the test reduces to the orthographic test with strict inequality, $a_E/b_E<a/c$.
\end{enumerate}
\end{proposition}

\begin{figure}[htbp]
\centering\includegraphics[width=0.55\linewidth]{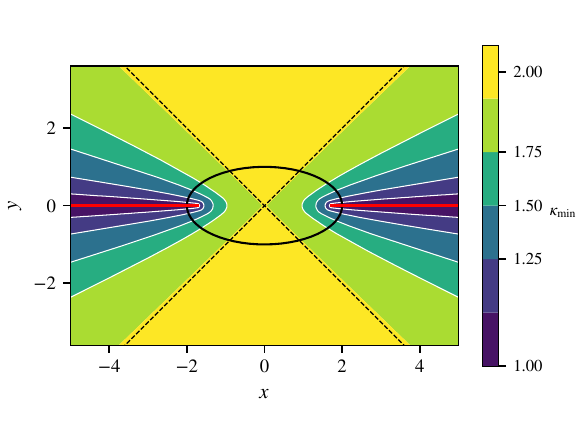}
\caption{$\kmin$ as a function of the principal point $(x,y)$, in the frame of an ellipse with semi-axes $2$ and $1$ (black). Where $|x|\le|y|$, outside the long-axis wedge bounded by the dashed diagonals, $\kmin=\kappa_\infty=2$ and no nonempty $I(\rho)$ is bounded above (\cref{prop:wedge}). Inside the wedge $\kmin<2$, and it drops to $1$ on the long axis line beyond the foci (red).}\label{fig:kmap}
\end{figure}

\Cref{prop:wedge} reflects foreshortening. For large $f$ the cone axis is tilted from the optical axis toward the center of $E$ by about $d_E/f$, so a cross-section perpendicular to it is approximately $E$ foreshortened along the line from $p$ to the center of $E$: rounder than $E$ if that line is nearer the long axis, more elongated if it is nearer the short axis. On the diagonals $|x|=|y|$ the $f^{-2}$ term vanishes, but still $\kappa>\kappa_\infty$ for every $f$ (proof of \cref{prop:wedge}; see also \cref{fig:kmap}).

\begin{example}\label{ex:ellipse}
Take the ellipse with center $(320,140)$~px, semi-axes $60$ and $38$~px, and long axis at $0.5$~rad to the $x$-axis, with $p$ at the origin (\cref{fig:gallery}(c)). The direction from its center to $p$ is $5^\circ$ from the long axis, so $p$ lies in the long-axis wedge and $f$ can be bounded above (\cref{prop:wedge}). Here $\kmin\approx1.0863$, attained at $f^*\approx287.2$~px (computed by \cref{thm:kminform} below): the ellipse is not a sphere outline at any $f$, but at some $f$ it is an outline of every ellipsoid with $a/c\ge\kmin$. For $a/c=1.5$ the admissible focal lengths are $I(1.5)=[164.1,1051.4]$~px; for $a/c=2$ they are all $f\ge116.8$~px, unbounded above because $2>\kappa_\infty=60/38\approx1.579$.
\end{example}

\subsection{The minimizing focal length and \texorpdfstring{$\kmin$}{kmin} in closed form}\label{sec:choosef}
If $\kmin$ is attained at $f^*$, then $I(\kmin)=\{f^*\}$, and $I(\rho)$ shrinks to $f^*$ as $\rho$ decreases to $\kmin$. This subsection characterizes $f^*$ geometrically, bounds how fast $\kappa$ can vary with $f$, and gives $\kmin$ in closed form. Let $v_\pm$ be unit eigenvectors of $Q_f$ for $\mu_\pm$, the axes of a cross-section of the cone perpendicular to its axis, and let $\varphi_\pm\in[0^\circ,90^\circ]$ be their angles with the optical axis $e_3$, so that $\cos\varphi_\pm=|v_\pm\cdot e_3|$.

\begin{proposition}[the stationarity condition]\label{prop:fstar}
For every $p$ and every $f>0$ with $\mu_-<\mu_+$,
\[
\frac{d\ln\kappa}{d\ln f}=\cos^2\varphi_--\cos^2\varphi_+ .
\]
Hence $\kappa$ is stationary exactly when the optical axis is equally inclined to the two cross-section axes, and an attained $\kmin>1$ is attained at such an $f^*$.
\end{proposition}

For $p$ off-axis, the condition reads $g(\theta_2)=g(\theta_3)$, with $\theta_2,\theta_3$ the negative confocal coordinates of $O$ (\cref{sec:where}) and $g(\theta)=x^2\theta^2/(a_E^2+\theta)^2+y^2\theta^2/(b_E^2+\theta)^2$.

\begin{corollary}[Lipschitz bound]\label{cor:lip}
$\ln\kappa$ is $1$-Lipschitz in $\ln f$: $|\ln\kappa(f')-\ln\kappa(f)|\le|\ln(f'/f)|$ for all $f,f'>0$. Hence, if $\kmin$ is attained at $f^*$, then $[f^*\kmin/\rho,\,f^*\rho/\kmin]\subseteq I(\rho)$ for every $\rho\ge\kmin$; for \cref{ex:ellipse} and $\rho=1.5$ this is $[208.0,396.6]$~px.
\end{corollary}

\paragraph{An algebraic parametrization} To compute the endpoints of $I(\rho)$ and $\kmin$, we replace $f$ by an eigenvalue: the level set $\kappa^2=k$ then becomes the zero set of a polynomial, whose roots can be counted and computed. Let $p$ be off-axis, and let
\[
q(u)=\det(A_E-uI)=(u-\alpha_1)(u-\alpha_2),\qquad \pi(u)=\det\bigl(C-u\diag(1,1,0)\bigr).
\]
Then $\pi$ has exactly one root $r$ in $(\alpha_1,\alpha_2)$; put $J=(\alpha_1,r)$ and $h(u)=\pi(u)/(u\,q(u))$. As $f$ increases from $0$ to $\infty$, the smaller positive eigenvalue of $Q_f$ is $f^2u$, where $u$ decreases from $r$ to $\alpha_1$ and $f^2=h(u)$. If $\kappa(f)^2=k>1$, then $u$ is a root of
\[
R_k(u)=\frac{k\,\pi(u)\,q(ku)-\pi(ku)\,q(u)}{1-k},
\]
a polynomial of degree at most four in $u$ (proof of \cref{thm:shape}(i)).

\begin{proposition}[endpoints of $I(\rho)$]\label{prop:endpoints}
Let $p$ be off-axis and $\rho>1$. The endpoints of $I(\rho)$ in $(0,\infty)$ are exactly the values $f=\sqrt{h(u)}$ for the roots $u\in J$ of $R_{\rho^2}$.
\end{proposition}

\begin{theorem}[$\kmin$ as a root of a quartic]\label{thm:kminform}
Let $p$ be off-axis with $\kmin$ attained (\cref{thm:shape}(iii)), and let $\mathcal D(k)$ be the discriminant of $R_k(u)$ as a polynomial of formal degree $4$ in $u$.
\begin{enumerate}
\item[(i)] $\kmin^2$ is the largest root of $\mathcal D$ in $(1,\infty)$; $R_{\kmin^2}$ has a double root $u^*\in J$, and $(f^*)^2=h(u^*)$.
\item[(ii)] Scale $C$ so that $A_E$ has eigenvalues $a_E^{-2},b_E^{-2}$. Then
\[
\mathcal D(k)=(a_Eb_E)^{-22}k^6\,\Omega_{E,p}(k+1/k)
\]
for a quartic $\Omega_{E,p}(w)=\sum_{i=0}^4\omega_iw^i$ whose coefficients are polynomials in $A=a_E^2$, $B=b_E^2$ and $(x,y)$, listed in the supplementary materials (structure in \cref{app:quartic}). The leading coefficient is
\[
\omega_4=A^3B^3(A-B)^2\,|p-F_1|^2|p-F_2|^2,
\]
where $F_1,F_2$ are the foci of $E$.
\end{enumerate}
Hence $\kmin^2+\kmin^{-2}$ is the largest real root of $\Omega_{E,p}$, and $\kmin$ follows by solving a quadratic.
\end{theorem}

The quartic comes from a symmetry: the substitution $u\mapsto ku$, $k\mapsto1/k$ maps $R_k$ to itself, so $\mathcal D(1/k)=k^{-12}\mathcal D(k)$. Since $\mathcal D$ has degree $10$ and is divisible by $k^2$ (proof of \cref{thm:kminform}), $k^{-2}\mathcal D$ is palindromic of degree $8$, hence $k^4$ times a quartic in $w=k+1/k$.

For \cref{ex:ellipse} the theorem gives $\kmin=1.08626$ and $f^*=287.24$~px; the identity for $\mathcal D$ and the $\omega_i$ were derived and checked by exact symbolic computation (see the proof). In double precision the quartic loses digits near degenerate configurations: in a stress test with attained minima (supplementary script \texttt{stress\_quartic.py}), its largest root gave $\kmin$ to a relative error of about $10^{-12}$ in generic configurations but only $10^{-8}$ near a focus and $5\cdot10^{-7}$ for nearly circular $E$, while in $40$-digit arithmetic it agreed with direct minimization to $10^{-29}$; the refinement in \cref{alg:test} removes this loss. In use, the quartic thus plays two roles: comparing $\rho$ with $\kmin$ decides exactly whether any focal length is admissible, and the double root $u^*$ gives a starting value for $f^*$ that a one-dimensional search polishes; the characterization is exact, and only the refinement is numerical. The attainment hypothesis is needed: for $a_E=3.6$, $b_E=1.2$ and $p=(2.15,0.068)$ in the frame of $E$, the unattained $\kmin^2=\kappa_0^2=5.8233$ gives $\kmin^2+\kmin^{-2}=5.9950$, whereas the largest root of $\Omega_{E,p}$ is $5.8113$.

\subsection{The algorithm}
\Cref{alg:test} combines these results. With $f$ unknown, Steps 3--6 decide analytically, without minimization, whether $I(\rho)$ is empty, unbounded above or bounded; Step~7 and the endpoint rule then compute $\kmin$ and $f^*$ (\cref{thm:kminform}) and the endpoints (\cref{prop:endpoints}).

\begin{algorithm}[tbp]
\caption{The ratio test with known or unknown focal length.}\label{alg:test}
\small
\textbf{Input:} ellipse $E$, principal point $p$, hypothesis $\rho=a/c\ge1$; optionally $f$.\\
\textbf{Output:} for known $f$, whether $E$ is an outline at $f$; for unknown $f$, the interval $I(\rho)$.\\
\emph{Axis formula} (proof of \cref{thm:shape}(ii)), for $p$ on an axis line of $E$, or $E$ a circle and $p$ not its center: choose unit eigenvectors $e_1,e_2$ of $A_E$ with $e_j$ along the line through $p$ and the center of $E$, let $l$ be the other index, and write $\beta_E=\beta_je_j$ and $\Delta=\alpha_j\gamma_E-\beta_j^2$. The endpoints of $I(\rho)$ in $(0,\infty)$ are the $f>0$ with $f^2=(\gamma_E\nu-\Delta)/(\nu(\nu-\alpha_j))$ for $\nu\in\{\rho^2\alpha_l,\alpha_l/\rho^2\}$.\\
\emph{Endpoint rule (unknown $f$).} A nonempty $I(\rho)$ has lower end $0$ exactly when $p$ is inside $E$ and either $\rho>\kappa_0$, or $\rho=\kappa_0$ and $\kappa$ does not increase near $f=0$. Its endpoints in $(0,\infty)$ are given by the axis formula if $p$ is on an axis line of $E$ or $E$ is a circle, and by \cref{prop:endpoints} otherwise.
\begin{enumerate}
\item \emph{Preprocessing.} Rescale image coordinates so that $E$ has size of order $1$, and compute $C$ in coordinates centered at $p$ (signed so that $A_E\succ0$), $\kappa_\infty$, the position $(x,y)$ of $p$ in the frame of $E$, $d_E$ and $e_E$, and, if $p$ is inside $E$, $\kappa_0$ and $\beta_\pm$. Convert back to pixels at the end.
\item \emph{Known $f$.} Accept if and only if $\kappa(f)\le\rho$ (\cref{thm:ellipsoid}). With an error bound on $C$ (\cref{rem:certified}), $\rho$ is certainly rejected if $\kappa_{\mathrm{lo}}>\rho$, certainly consistent if $\kappa_{\mathrm{hi}}\le\rho$, and undecided otherwise.
\end{enumerate}
\emph{Unknown $f$:} apply the first of Steps 3--7 whose condition holds.
\begin{enumerate}\setcounter{enumi}{2}
\item If $\rho>\kappa_\infty$, or $\rho=\kappa_\infty$ and either $p$ is the center of $E$ or ($E$ is not a circle and $|x|>|y|$): $I(\rho)$ is nonempty and unbounded above (for $\rho=\kappa_\infty$ and $|x|>|y|$, since $\kappa<\kappa_\infty$ for large $f$ by \cref{prop:wedge}(i)). It is $(0,\infty)$ if $p$ is the center of $E$, and otherwise $(0,\infty)$ or $[f_{\min},\infty)$ as given by the endpoint rule.
\item If $E$ is a circle or $|x|\le|y|$: $I(\rho)=\emptyset$ (\cref{prop:kmin,prop:wedge}). From here on $\rho<\kappa_\infty$, so a nonempty $I(\rho)$ is bounded above.
\item If $p$ is on the long axis line of $E$: for $d_E>e_E$, $\kmin=1$ is attained at the $f$ of \cref{rem:special}; otherwise $\kmin=\kappa_0$ and $\kappa$ increases strictly. $I(\rho)=\emptyset$ if $\rho<\kmin$ or $\rho=\kmin=\kappa_0$.
\item If $p$ is inside $E$ and $\beta_+^2\le\beta_-^2$: $\kappa$ increases strictly from $\kappa_0$ (\cref{thm:shape}(iii)), so $I(\rho)=\emptyset$ for $\rho\le\kappa_0$ and $I(\rho)=(0,f_{\max}]$ otherwise, with $f_{\max}$ given by \cref{prop:endpoints}.
\item Otherwise $\kmin$ is attained. Compute the largest real root $w$ of $\Omega_{E,p}$, $\kmin$ from $\kmin^2+\kmin^{-2}=w$, the common root $u^*\in J$ of $R_{\kmin^2}$ and $\partial_uR_{\kmin^2}$, and $f^*=\sqrt{h(u^*)}$. Refine $f^*$ by golden-section search for the minimum of $\ln\kappa$ over $\ln f$ on the bracket $[f^*/m,mf^*]$, $m=1.1$, enlarged by factors of $m$ until $\kappa$ at both ends exceeds $\kappa$ at $f^*$; it converges by \cref{thm:shape}(ii), and it restores the digits that the quartic loses in floating point near degenerate configurations (\cref{sec:choosef}). $I(\rho)=\emptyset$ if $\rho<\kmin$. Otherwise the endpoints of $I(\rho)$ are given by the endpoint rule ($f=\sqrt{h(u)}$ for the roots $u\in J$ of $R_{\rho^2}$, and $0$ if applicable), and $I(\rho)$ contains the guaranteed inner interval $[f^*\kmin/\rho,\,f^*\rho/\kmin]$ (\cref{cor:lip}).
\end{enumerate}
\end{algorithm}

\section{Limits and extensions}\label{sec:practical}

\subsection{What a single ellipse can tell}

\begin{corollary}[a single ellipse bounds elongation only from below]\label{cor:nested}
Fix $p$, and let $\mathcal E_\rho$ be the set of ellipses that are outlines, for some $f>0$, of an ellipsoid with $a/c=\rho$. Then $E\in\mathcal E_\rho$ if $\kmin(E,p)<\rho$, $E\notin\mathcal E_\rho$ if $\kmin(E,p)>\rho$, if $\kmin(E,p)=\rho$ then $E\in\mathcal E_\rho$ exactly when the infimum is attained, and $\mathcal E_{\rho_1}\subseteq\mathcal E_{\rho_2}$ whenever $\rho_1<\rho_2$.
\end{corollary}

\begin{proof}
The first three claims are \cref{prop:kmin}; the inclusion follows from \cref{thm:ellipsoid}, since $\kappa(f)\le\rho_1$ implies $\kappa(f)\le\rho_2$.
\end{proof}

So a single ellipse can identify a shape among hypotheses $\rho_1<\dots<\rho_n$ only if that shape is the most elongated one, $\rho_n$, and $E\in\mathcal E_{\rho_n}\setminus\mathcal E_{\rho_{n-1}}$; otherwise every hypothesis from the roundest consistent one up to $\rho_n$ remains possible.

\paragraph{The principal point is essential} With $p$ unknown, one ellipse rules out no ellipsoid, even with $f$ known: a suitable $p$ on the long axis line beyond a focus (\cref{lem:sphere}), or the center of a circle, makes $E$ a sphere outline, hence an outline of every ellipsoid. If $p$ is only known to lie in a region $\Pset$, the test with $f$ unknown becomes $\inf_{p\in\Pset}\kmin(E,p)\le a/c$ (with equality only if attained), which is vacuous once $\Pset$ meets the long axis line beyond a focus. In \cref{ex:ellipse} that line passes $30.6$~px from $p$, whereas with $f=800$~px known the sphere-outline principal point is $631$~px away.

\paragraph{Several ellipsoids in one image} Each object gives inequalities, so several ellipsoids generically bound $f$ and $p$ but, unlike spheres, do not determine them. Ellipsoids of known shape seen by one camera with known $p$ admit a common $f$ exactly when their intervals $I_j(\rho_j)$ intersect, and by Helly's theorem on the line it suffices that they intersect pairwise; this uses that $p$ is known, since for unknown $p$ the regions of camera centers are not convex. The same holds for several images taken with one unknown $f$; with $p$ unknown, the admissible camera centers are the intersection of the regions of \cref{prop:confocal}. The supplementary materials give a numerical example.

\paragraph{Several views of one ellipsoid} With unknown relative poses the views are independent, so they are compatible with ratio $a/c$ exactly when each passes the ratio test, $\max_i\kappa_i\le a/c$, and $b$ stays unobservable. With known relative poses, two outlines of a quadric must satisfy two epipolar tangency conditions, which leave a one-parameter family of quadrics \cite{CrossZisserman1998}; when this family contains an ellipsoid with semi-axes in the ratio $a:b:c$, now depending on $b$, is an open problem.

\subsection{Noisy ellipses and general calibration}

\begin{remark}[the error-bound test]\label{rem:certified}
Rescale image coordinates to size of order $1$ by a factor $s$ (for example, $s=1000$~px): divide $f$ and $p$ by $s$ and replace $C$ by $\diag(s,s,1)C\diag(s,s,1)$. Let the fitted conic matrix $\hat C$ satisfy $\|\hat C-C\|_2\le\varepsilon$, with both matrices normalized to unit Frobenius norm and signed as in \cref{sec:setting}. Then $K_f\hat CK_f$, with positive eigenvalues $\hat\mu_-\le\hat\mu_+$, differs from $Q_f$ by at most $\delta(f)=\max(f,1)^2\varepsilon$ in spectral norm, and by Weyl's inequality so does each sorted eigenvalue. If $\hat\mu_->\delta$,
\[
\kappa_{\mathrm{lo}}=\sqrt{\frac{\hat\mu_+-\delta}{\hat\mu_-+\delta}}\le\kappa\le\sqrt{\frac{\hat\mu_++\delta}{\hat\mu_--\delta}}=\kappa_{\mathrm{hi}}.
\]
At a given $f$, we call $a/c$ \emph{certainly rejected} if $\kappa_{\mathrm{lo}}>a/c$, \emph{certainly consistent} if $\kappa_{\mathrm{hi}}\le a/c$, and \emph{undecided} otherwise. The guarantee is deterministic when $\varepsilon$ is a true bound. When $\varepsilon$ is estimated, for example by bootstrap, it is only statistical: it holds as often as the estimate does and misses systematic errors such as a biased fit. The point test ignores the error altogether. Weyl's bound is worst case over all perturbations of norm $\varepsilon$; when the covariance of the fit is known, the first-order changes $v_\pm\T K_f(\hat C-C)K_fv_\pm$ of $\mu_\pm$, with $v_\pm$ the unit eigenvectors, give tighter but only statistical intervals. The bound is useless near $f=0$, where $\mu_-=O(f^2)$, and in pixel units: for \cref{ex:ellipse} at $f^*$ it tolerates $\varepsilon$ only up to about $2\cdot10^{-5}$, against $0.07$ after rescaling with $s=1000$~px. An error in the principal point can be absorbed in the same way: moving $p$ by $\Delta p$ (in rescaled units, $|\Delta p|=\eta$) replaces $C$ by $T\T CT$ with $T=\bigl(\begin{smallmatrix}I&\Delta p\\0&1\end{smallmatrix}\bigr)$, and $\|T\T CT-C\|_F\le(2\eta+\eta^2)\|C\|_F$; after renormalization the change is at most $2(2\eta+\eta^2)$, which can be added to $\varepsilon$.
\end{remark}

For all focal lengths above a chosen one, the bound can be made uniform in $f$.

\begin{proposition}[a uniform error bound]\label{prop:uniform}
Rescale image coordinates by $s>0$, and let $C$ and $\hat C$ be real ellipses, normalized as in \cref{rem:certified}, with $\|\hat C-C\|_2\le\varepsilon<\hat\alpha_1$, where $\hat\alpha_1$ is the smaller eigenvalue of $\hat A_E$. Put
\[
\tau_\varepsilon=\tfrac12\ln\frac{\hat\alpha_1+\varepsilon}{\hat\alpha_1-\varepsilon}.
\]
Then $|\ln\kappa(f)-\ln\hat\kappa(f)|\le\tau_\varepsilon$ for every $f\ge s$. Consequently the infima of $\kappa$ and $\hat\kappa$ over $f\ge s$ differ by a factor at most $e^{\tau_\varepsilon}$, and $\hat I(e^{-\tau_\varepsilon}\rho)\cap[s,\infty)\subseteq I(\rho)\cap[s,\infty)\subseteq\hat I(e^{\tau_\varepsilon}\rho)$, where $\hat I$ refers to $\hat C$.
\end{proposition}

\begin{proof}
In rescaled units $f\ge1$, so $\|K_f(\hat C-C)K_f\|_2\le f^2\varepsilon$, and by Weyl's inequality $|\mu_\pm-\hat\mu_\pm|\le f^2\varepsilon$. Cauchy interlacing of $K_f\hat CK_f$ with its block $f^2\hat A_E$ gives $\hat\mu_+\ge\hat\mu_-\ge f^2\hat\alpha_1$. Let $\xi=\varepsilon/\hat\alpha_1<1$. Then $|\mu_\pm-\hat\mu_\pm|\le f^2\varepsilon\le\xi\hat\mu_\pm$, so $1-\xi\le\mu_+/\hat\mu_+\le1+\xi$ and $1/(1+\xi)\le\hat\mu_-/\mu_-\le1/(1-\xi)$, and $\kappa^2/\hat\kappa^2=(\mu_+/\hat\mu_+)(\hat\mu_-/\mu_-)$ lies between $(1-\xi)/(1+\xi)$ and $(1+\xi)/(1-\xi)$. The other claims follow.
\end{proof}

So, with $s$ a lower bound on the focal length, $a/c$ is certainly rejected for every $f\ge s$ if $\inf_{f\ge s}\hat\kappa>e^{\tau_\varepsilon} a/c$, and certainly consistent for some $f\ge s$ if $\inf_{f\ge s}\hat\kappa<e^{-\tau_\varepsilon}a/c$; by \cref{thm:shape}, the infimum is $\hat\kappa_{\min}$ or $\hat\kappa(s)$. Here $\tau_\varepsilon\approx\varepsilon/\hat\alpha_1$, the relative error of the conic measured against its quadratic part, and a principal-point error can be added to $\varepsilon$ as in \cref{rem:certified}. In practice, running \cref{alg:test} on $\hat C$ with the hypotheses $e^{-\tau_\varepsilon}\rho$ and $e^{\tau_\varepsilon}\rho$ gives inner and outer bounds on $I(\rho)\cap[s,\infty)$.

\begin{remark}[general calibration]\label{rem:generalK}
A known calibration reduces to the case treated so far: for an upper-triangular $K$ and $W=K^{-\mathsf T}K^{-1}$, the cone of rays $K\T CK$ has the generalized eigenvalues of the pencil $(C,W)$, as in calibration from spheres \cite{AgrawalDavis2003,YingZha2006,ZhangWongZhang2007}, and the test is $\sqrt{\zeta_+/\zeta_-}\le a/c$ for the two eigenvalues $\zeta_\pm$ of equal sign; for $K=K_0\diag(f,f,1)$ with $K_0$ known, \cref{sec:focal} applies to $K_0\T CK_0$. If the aspect ratio and skew are unknown as well, the test is vacuous: for every ellipse, some $K$ with principal point $p$ makes the cone of rays circular (construction in the supplementary materials).
\end{remark}

\section{Experiments}\label{sec:numerics}
\subsection{Simulations}
We ask how noise in the fitted ellipse changes the decisions of the ratio test, with $f$ known and unknown, and what the error-bound test costs in power; details are in the supplementary materials.

\paragraph{Setup} For each true ratio $a/c\in\{1,1.05,1.1,1.2,1.5,2,3\}$ we drew 500 scenes ($c=1$, $b$ log-uniform in $[c,a]$, random orientation, depth $8$ to $20$, center projecting to a random point of a $1000\times1000$~px image, $f=800$~px, $p$ at the image center). We sampled 120 outline points, added Gaussian noise of $\sigma=0.5$~px and fitted a conic by the direct least-squares fit \cite{Fitzgibbon1999,HalirFlusser1998}; $\hat\kappa$ and $\hat\kappa_{\min}$ denote $\kappa$ and $\kmin$ of the fitted conic, computed on a logarithmic grid of $900$ focal lengths refined by golden-section and bisection steps, which \cref{thm:shape} makes reliable. The \emph{point test} accepts a hypothesis $\rho$ for $a/c$ if $\hat\kappa(f)\le\rho$ ($f$ known) or $\hat\kappa_{\min}\le\rho$ ($f$ unknown). Hypotheses more elongated than the truth are consistent with every exact outline, so we test only rounder ones, and we report each rejection rate next to its \emph{noise-free reference}, the rate for the exact outlines.

\begin{table}[tbp]
\centering\footnotesize\setlength{\tabcolsep}{3pt}
\caption{Point test on noisy outlines: percent of the 500 outlines per true $a/c$ (Wilson $95\%$ half-widths at most $4.4$ percentage points), with the noise-free reference in parentheses; in each column pair, $f$ known and unknown. ``Accept true'': the hypothesis equals the true $a/c$. ``--'' in a rejection column: the hypothesis is not rounder than the truth. Last column: percent of outlines with $I(a/c)$ bounded above, among those with $I(a/c)$ nonempty (``--'' for spheres, whose $I(1)$ is empty because $\hat\kappa>1$ almost surely).}\label{tab:e1}
\begin{tabular}{ccccccccc}
\toprule
 & \multicolumn{2}{c}{accept true} & \multicolumn{2}{c}{reject hypothesis $1.05$} & \multicolumn{2}{c}{reject hypothesis $1.2$} & $I(a/c)$\\
\cmidrule(lr){2-3}\cmidrule(lr){4-5}\cmidrule(lr){6-7}
true $a/c$ & known & unknown & known & unknown & known & unknown & bdd.\ above\\\midrule
1 & 0.0 & 0.0 & -- & -- & -- & -- & --\\
1.05 & 96.2 & 99.8 & -- & -- & -- & -- & 76.4\\
1.1 & 99.6 & 100.0 & 65.8 (65.0) & 34.2 (34.4) & -- & -- & 51.4\\
1.2 & 99.6 & 99.8 & 87.2 (87.6) & 66.6 (66.2) & -- & -- & 24.6\\
1.5 & 99.8 & 100.0 & 97.0 (97.0) & 88.8 (88.6) & 71.2 (71.0) & 49.4 (49.4) & 8.6\\
2 & 99.8 & 100.0 & 98.8 (98.6) & 91.4 (91.4) & 87.0 (87.0) & 72.6 (72.4) & 5.2\\
3 & 100.0 & 100.0 & 99.0 (99.0) & 96.4 (96.4) & 95.2 (95.2) & 88.2 (88.2) & 3.0\\
\bottomrule
\end{tabular}

\end{table}

\paragraph{Decisions (\cref{tab:e1})} Noise rarely changes a decision, and then only near the bound. With $f$ known, all 25 false rejections among the 3000 outlines with $a/c>1$ had relative margin $(a/c-\kappa)/(a/c)<1\%$ (13 of the 41 below $0.1\%$, none of the 2676 above $1\%$); with $f$ unknown, the true shape was accepted for at least $99.8\%$. Rejection of rounder hypotheses matched the noise-free references within one percentage point, so it is limited by geometry, not noise; it is weaker with $f$ unknown. Upper bounds on $f$ come mainly from nearly round objects ($76\%$ of the outlines with $a/c=1.05$, $3\%$ with $a/c=3$) and, as \cref{prop:wedge} predicts, only from principal points in the long-axis wedge (all 845 bounded intervals, none of the 846 nonempty intervals outside the wedge), and the median ratio of the ends of a bounded $I(a/c)$ grew from $1.65$ to $15$ (Table~SM7). The minimizer $f^*$ is not an estimate of $f$ (median relative error $12$ to $65\%$ for $a/c\ge1.05$). Since $\hat\kappa>1$ almost surely, the point test rejects every true sphere; near-spherical hypotheses need the error-bound test.

\paragraph{The error-bound test} We took $\varepsilon$ as the $95$th percentile of $\|\hat C^{(i)}-\hat C\|_2$ over $100$ bootstrap refits (Table~SM1); it bounded the true error for $93$ to $96\%$ of the outlines, and whenever it did, no true shape was certainly rejected, as \cref{rem:certified} guarantees. True spheres were left undecided (two exceptions, with too small $\varepsilon$). Certainty costs little power: $\rho=1.2$ was certainly rejected for $69$, $86$ and $95\%$ of the outlines with $a/c=1.5$, $2$ and $3$ with $f$ known (point test: $71$, $87$, $95\%$) and, with $f$ unknown, by the uniform bound of \cref{prop:uniform} with $s=200$~px (median $\tau_\varepsilon=0.011$), for $47$, $74$ and $88\%$ (point test over all $f>0$: $49$, $73$, $88\%$); the uniform bound never certainly rejected a true shape.

\paragraph{Robustness and baselines} Noise up to $2$~px, $30$ to $240$ points, half an outline, $5\%$ outliers with RANSAC \cite{FischlerBolles1981}, and $f=400$ or $1600$~px kept acceptance at $99.0$ to $100\%$ (Table~SM2). The risks are a quarter arc, which biases the fit (true shape accepted for $60\%$ with $f$ known), uncorrected distortion and nonellipsoidal objects ($95.8$ to $98.5\%$), and an error in $p$, which weakens rejection with $f$ unknown (\cref{sec:practical}). A grid scan of \cite[Thm.~1]{Gaudilliere2023} agreed with the ratio test in 1795 of 1800 decisions and pose fitting with a $\chi^2$ test in 1776 (Table~SM3); a decision took $0.016$~ms, and $\kmin$ from the quartic $0.26$~ms.

\FloatBarrier
\subsection{Real images: Saturn's moons}\label{sec:cassini}
\paragraph{Data and limb extraction} We used the 50 images of Saturn's moons selected by \cite{DanasRiveraPeck2023} from the Cassini narrow-angle camera (NAC) \cite{Porco2004}, calibrated with CISSCAL~4.0 \cite{Knowles2020} (\cite{COISSdata}; list in Table~SM6), excluding five with edge points within $10$~px of the frame. Here $f=166{,}953$~px, so perspective is negligible: these images check the test with known $f$, not the results for unknown $f$. The shape kernel \texttt{cpck15Dec2017.tpc} \cite{Thomas2010,Acton2018} gives $a/c$ from $1.006$ (Rhea, Dione) to $1.088$ (Mimas), and \cref{lem:cone} with the spacecraft position gives the predicted cone ratio $\kappa_{\mathrm{pred}}$. Only the sunlit part of the \emph{limb}, the moon's outline, is visible; the rest of the bright boundary is the \emph{terminator}, which is not an outline. We kept subpixel edge points within $85^\circ$ of the projected Sun direction (a half-angle chosen after inspecting the results; Table~SM5) and fitted a conic with RANSAC (median 356 points, $47\%$ of the outline; thresholds in Table~SM4).

\begin{table}[tbp]
\centering\footnotesize\setlength{\tabcolsep}{2.5pt}
\caption{Cassini NAC limbs ($n$ images used, 45 in all). $\kappa_{\mathrm{pred}}$: cone ratio of the ellipsoid model in the PDS geometry. Margin: $a/c-\kappa_{\mathrm{pred}}\ge0$, the distance of the true view from the bound. Point: true shape accepted by the point test ($\hat\kappa\le a/c$). Bootstrap: the error-bound interval $[\kappa_{\mathrm{lo}},\kappa_{\mathrm{hi}}]$ with bootstrap $\varepsilon$ contains $\kappa_{\mathrm{pred}}$. Allowance $\delta_\kappa$: $90$th percentile of $|\hat\kappa-\kappa_{\mathrm{pred}}|$ on the other moons ($0.014$ for Rhea and Dione, $0.008$ to $0.012$ for the others); the columns give the number of images for which, with the interval $[\hat\kappa-\delta_\kappa,\hat\kappa+\delta_\kappa]$, the true shape is consistent / undecided / rejected, and for which the sphere is rejected ($\hat\kappa-\delta_\kappa>1$).}\label{tab:cassini}
\begin{tabular}{lcccccccc}
\toprule
 & & & median & median & point & bootstrap & \multicolumn{2}{c}{with allowance $\delta_\kappa$}\\
\cmidrule(lr){8-9}
moon & $n$ & $a/c$ & margin & $|\hat\kappa-\kappa_{\mathrm{pred}}|$ & accepts & covers $\kappa_{\mathrm{pred}}$ & cons./und./rej. & sphere rej.\\\midrule
Rhea & 10 & 1.006 & 0.005 & 0.001 & 7/10 & 5/10 & 0/10/0 & 0/10\\
Dione & 7 & 1.006 & 0.003 & 0.002 & 5/7 & 3/7 & 0/7/0 & 0/7\\
Tethys & 11 & 1.024 & 0.010 & 0.008 & 9/11 & 3/11 & 7/3/1 & 10/11\\
Enceladus & 11 & 1.033 & 0.009 & 0.004 & 9/11 & 2/11 & 3/7/1 & 9/11\\
Iapetus & 1 & 1.049 & 0.002 & 0.001 & 1/1 & 1/1 & 0/1/0 & 1/1\\
Mimas & 5 & 1.088 & 0.030 & 0.008 & 5/5 & 3/5 & 4/1/0 & 5/5\\
\midrule
all & 45 & -- & 0.005 & 0.004 & 36/45 & 17/45 & 14/29/2 & 25/45\\
\bottomrule
\end{tabular}

\end{table}

\begin{figure}[tbp]
\centering\includegraphics[width=\linewidth]{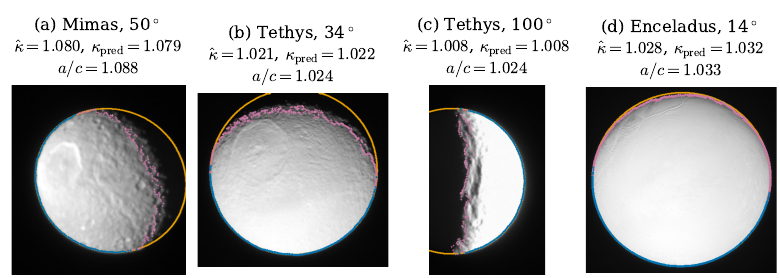}
\caption{Limb fits on four of the 45 Cassini NAC images (cropped). Blue dots: limb points used (within $85^\circ$ of the projected Sun direction); magenta crosses: other edge points (terminator, or limb beyond the $85^\circ$ cut); orange: fitted outline. Above each panel: moon, phase angle, measured cone ratio $\hat\kappa$, predicted cone ratio $\kappa_{\mathrm{pred}}$, and the moon's $a/c$.}\label{fig:cassini}
\end{figure}

\paragraph{Results (\cref{tab:cassini}, \cref{fig:cassini})} Our limb extraction has a systematic error of about $1\%$ in $\kappa$, larger than the noise: the median $|\hat\kappa-\kappa_{\mathrm{pred}}|$ is $0.004$ ($90\%$ below $0.012$), against $0.011$ for a sphere model, but a constant per moon fits as well, and the images of one moon are not independent, so the counts describe six moons rather than 45 trials. The views lie close to the bound (median margin $0.005$). The point test accepted the true shape for 36 of 45 images, all rejections having margin below $0.02$. The bootstrap error bound misses the systematic error (it covered $\kappa_{\mathrm{pred}}$ for only 17 images), so we widened the interval by an allowance $\delta_\kappa$ calibrated by leaving one moon out: for the images of each moon, $\delta_\kappa$ is the $90$th percentile of $|\hat\kappa-\kappa_{\mathrm{pred}}|$ over the images of the other five moons; this allowance is specific to our pipeline and outside the guarantee of \cref{rem:certified}. With it, the true shape was rejected for 2 of 45 images and the sphere for 25, never for Rhea or Dione, whose $a/c-1=0.006$ is below $\delta_\kappa$. The counts are stable under the half-angle and the random seed but depend on the shape model (31 of 45 accepted with \texttt{pck00011.tpc}).

\paragraph{Reproducibility} The scripts in \texttt{experiments/} reproduce every number of this section with fixed seeds, and \texttt{sanity\_checks.py} checks \cref{lem:cone,prop:confocal} and \cref{thm:shape}(i) numerically (supplementary materials); all code is at \url{https://github.com/michael-werman/realizability-code}.

\section{Conclusions}\label{sec:conclusions}
The ratio test reduces realizability to one eigenvalue inequality, and its analysis in $f$ gives an interval, a closed-form minimum and a simple condition for an upper bound. In practice, near-spherical hypotheses need the error-bound test (\cref{rem:certified,prop:uniform}), and the main risks are short arcs, lens distortion, an uncertain principal point and systematic limb errors (about $1\%$ in $\kappa$ on the Cassini images). Real images with strong perspective are still needed to test the results for unknown $f$. Open questions are how fast the bounds tighten with several objects and the compatibility of several views with known relative poses (\cref{sec:practical}).

\section*{Acknowledgments}
The author used Claude (Anthropic; Claude Opus 5.5) as an assistant to edit the text, check proofs, search the literature, and write the code for the experiments and figures. All statements and proofs were checked by the author, and every numerical result is produced by the published code. The Cassini images and kernels were obtained from the PDS Ring-Moon Systems Node and NASA's Navigation and Ancillary Information Facility. The author assumes responsibility for all content.

\bibliographystyle{siamplain}
\bibliography{refs}

@inproceedings{AgrawalDavis2003,
  author = {M. Agrawal and L. S. Davis},
  title = {Camera calibration using spheres: A semi-definite programming approach},
  booktitle = {Proc. Ninth IEEE International Conference on Computer Vision (ICCV)},
  volume = {2}, year = {2003}, pages = {782--789},
  doi = {10.1109/ICCV.2003.1238428}}

@article{Binney1985,
  author = {J. Binney},
  title = {Testing for triaxiality with kinematic data},
  journal = {Mon. Not. R. Astron. Soc.}, volume = {212}, number = {4}, year = {1985}, pages = {767--781},
  doi = {10.1093/mnras/212.4.767}}

@article{Christian2021,
  author = {J. A. Christian},
  title = {A tutorial on horizon-based optical navigation and attitude determination with space imaging systems},
  journal = {IEEE Access}, volume = {9}, year = {2021}, pages = {19819--19853},
  doi = {10.1109/ACCESS.2021.3051914}}

@inproceedings{CrossZisserman1998,
  author = {G. Cross and A. Zisserman},
  title = {Quadric reconstruction from dual-space geometry},
  booktitle = {Proc. Sixth International Conference on Computer Vision (ICCV)}, year = {1998}, pages = {25--31},
  doi = {10.1109/ICCV.1998.710697}}

@misc{DanasRiveraPeck2023,
  author = {K. R. Danas Rivera and M. A. Peck},
  title = {Camera calibration from a single imaged ellipsoid: A moon calibration algorithm},
  year = {2023}, howpublished = {Preprint, arXiv:2307.00689 [cs.CV]}}

@inproceedings{Daucher1994,
  author = {N. Daucher and M. Dhome and J. T. Laprest\'e},
  title = {Camera calibration from spheres images},
  booktitle = {Computer Vision -- ECCV '94}, editor = {J.-O. Eklundh},
  series = {Lecture Notes in Comput. Sci.}, volume = {800},
  publisher = {Springer}, address = {Berlin, Heidelberg}, year = {1994}, pages = {447--454},
  doi = {10.1007/3-540-57956-7_49}}

@article{Gaudilliere2023,
  author = {V. Gaudilli\`ere and G. Simon and M.-O. Berger},
  title = {Perspective-1-Ellipsoid: Formulation, analysis and solutions of the camera pose estimation problem from one ellipse-ellipsoid correspondence},
  journal = {Int. J. Comput. Vis.}, volume = {131}, number = {9}, year = {2023}, pages = {2446--2470},
  doi = {10.1007/s11263-023-01794-x}, note = {Extended version: arXiv:2208.12513v3}}

@article{Golub1973,
  author = {G. H. Golub},
  title = {Some modified matrix eigenvalue problems},
  journal = {SIAM Rev.}, volume = {15}, number = {2}, year = {1973}, pages = {318--334},
  doi = {10.1137/1015032}}

@book{HartleyZisserman2004,
  author = {R. Hartley and A. Zisserman},
  title = {Multiple View Geometry in Computer Vision}, edition = {2nd},
  publisher = {Cambridge University Press}, address = {Cambridge}, year = {2004},
  doi = {10.1017/CBO9780511811685}}

@article{Nicholson2019,
  author = {L. Nicholson and M. Milford and N. S\"underhauf},
  title = {{QuadricSLAM}: Dual quadrics from object detections as landmarks in object-oriented {SLAM}},
  journal = {IEEE Robot. Autom. Lett.}, volume = {4}, number = {1}, year = {2019}, pages = {1--8},
  doi = {10.1109/LRA.2018.2866205}}

@inproceedings{Penne2015,
  author = {R. Penne and B. Ribbens and L. Mertens and P. Levrie},
  title = {What does one image of one ball tell us about the focal length?},
  booktitle = {Advanced Concepts for Intelligent Vision Systems (ACIVS)},
  editor = {S. Battiato and J. Blanc-Talon and G. Gallo and W. Philips and D. Popescu and P. Scheunders},
  series = {Lecture Notes in Comput. Sci.}, volume = {9386},
  publisher = {Springer}, address = {Cham}, year = {2015}, pages = {501--509},
  doi = {10.1007/978-3-319-25903-1_43}}

@article{Rubino2018,
  author = {C. Rubino and M. Crocco and A. {Del Bue}},
  title = {{3D} object localisation from multi-view image detections},
  journal = {IEEE Trans. Pattern Anal. Mach. Intell.}, volume = {40}, number = {6}, year = {2018}, pages = {1281--1294},
  doi = {10.1109/TPAMI.2017.2701373}}

@book{Salmon1882,
  author = {G. Salmon},
  title = {A Treatise on the Analytic Geometry of Three Dimensions}, edition = {4th},
  publisher = {Hodges, Figgis}, address = {Dublin}, year = {1882}}

@book{Smith1907,
  author = {C. Smith},
  title = {An Elementary Treatise on Solid Geometry}, edition = {2nd},
  publisher = {Macmillan}, address = {London}, year = {1907},
  note = {First edition 1884; 11th impression of the 2nd edition}}

@article{Stark1977,
  author = {A. A. Stark},
  title = {Triaxial models of the bulge of {M31}},
  journal = {Astrophys. J.}, volume = {213}, year = {1977}, pages = {368--373},
  doi = {10.1086/155164}}

@article{WokesPalmer2010,
  author = {D. S. Wokes and P. L. Palmer},
  title = {Perspective reconstruction of a spheroid from an image plane ellipse},
  journal = {Int. J. Comput. Vis.}, volume = {90}, number = {3}, year = {2010}, pages = {369--379},
  doi = {10.1007/s11263-010-0368-0}}

@article{YingZha2006,
  author = {X. Ying and H. Zha},
  title = {Geometric interpretations of the relation between the image of the absolute conic and sphere images},
  journal = {IEEE Trans. Pattern Anal. Mach. Intell.}, volume = {28}, number = {12}, year = {2006}, pages = {2031--2036},
  doi = {10.1109/TPAMI.2006.245}}

@article{ZhangWongZhang2007,
  author = {H. Zhang and K.-Y. K. Wong and G. Zhang},
  title = {Camera calibration from images of spheres},
  journal = {IEEE Trans. Pattern Anal. Mach. Intell.}, volume = {29}, number = {3}, year = {2007}, pages = {499--502},
  doi = {10.1109/TPAMI.2007.45}}

@article{FerriMangiliViano1993,
  author = {M. Ferri and F. Mangili and G. Viano},
  title = {Projective pose estimation of linear and quadratic primitives in monocular computer vision},
  journal = {CVGIP: Image Underst.}, volume = {58}, number = {1}, year = {1993}, pages = {66--84},
  doi = {10.1006/cviu.1993.1033}}

@article{Forsyth1996,
  author = {D. A. Forsyth},
  title = {Recognizing algebraic surfaces from their outlines},
  journal = {Int. J. Comput. Vis.}, volume = {18}, number = {1}, year = {1996}, pages = {21--40}}

@article{GaudilliereP2E2020,
  author = {V. Gaudilli\`ere and G. Simon and M.-O. Berger},
  title = {Perspective-2-Ellipsoid: Bridging the gap between object detections and 6-{DoF} camera pose},
  journal = {IEEE Robot. Autom. Lett.}, volume = {5}, number = {4}, year = {2020}, pages = {5189--5196},
  doi = {10.1109/LRA.2020.3005387}}

@book{GlaeserStachelOdehnal2016,
  author = {G. Glaeser and H. Stachel and B. Odehnal},
  title = {The Universe of Conics},
  publisher = {Springer}, address = {Berlin, Heidelberg}, year = {2016},
  doi = {10.1007/978-3-662-45450-3}}

@inproceedings{HanWongTan2016,
  author = {K. Han and K.-Y. K. Wong and X. Tan},
  title = {Single view {3D} reconstruction under an uncalibrated camera and an unknown mirror sphere},
  booktitle = {Proc. Fourth International Conference on 3D Vision (3DV)}, year = {2016}, pages = {408--416},
  doi = {10.1109/3DV.2016.50}}

@article{Ma1993,
  author = {S. D. Ma},
  title = {Conics-based stereo, motion estimation, and pose determination},
  journal = {Int. J. Comput. Vis.}, volume = {10}, number = {1}, year = {1993}, pages = {7--25},
  doi = {10.1007/BF01440844}}

@article{ZinsSimonBerger2022,
  author = {M. Zins and G. Simon and M.-O. Berger},
  title = {Object-based visual camera pose estimation from ellipsoidal model and {3D}-aware ellipse prediction},
  journal = {Int. J. Comput. Vis.}, volume = {130}, number = {4}, year = {2022}, pages = {1107--1126},
  doi = {10.1007/s11263-022-01585-w}}

@article{Fitzgibbon1999,
  author = {A. Fitzgibbon and M. Pilu and R. B. Fisher},
  title = {Direct least square fitting of ellipses},
  journal = {IEEE Trans. Pattern Anal. Mach. Intell.}, volume = {21}, number = {5}, year = {1999}, pages = {476--480},
  doi = {10.1109/34.765658}}

@inproceedings{HalirFlusser1998,
  author = {R. Hal\'{\i}\v{r} and J. Flusser},
  title = {Numerically stable direct least squares fitting of ellipses},
  booktitle = {Proc. 6th International Conference in Central Europe on Computer Graphics and Visualization (WSCG)}, year = {1998}, pages = {125--132}}

@article{FischlerBolles1981,
  author = {M. A. Fischler and R. C. Bolles},
  title = {Random sample consensus: A paradigm for model fitting with applications to image analysis and automated cartography},
  journal = {Commun. ACM}, volume = {24}, number = {6}, year = {1981}, pages = {381--395},
  doi = {10.1145/358669.358692}}

@article{Porco2004,
  author = {C. C. Porco and others},
  title = {Cassini imaging science: Instrument characteristics and anticipated scientific investigations at {S}aturn},
  journal = {Space Sci. Rev.}, volume = {115}, year = {2004}, pages = {363--497},
  doi = {10.1007/s11214-004-1456-7}}

@article{Thomas2010,
  author = {P. C. Thomas},
  title = {Sizes, shapes, and derived properties of the saturnian satellites after the {C}assini nominal mission},
  journal = {Icarus}, volume = {208}, number = {1}, year = {2010}, pages = {395--401},
  doi = {10.1016/j.icarus.2010.01.025}}

@misc{COISSdata,
  author = {{Cassini Imaging Science Subsystem Team}},
  title = {Cassini {ISS} {S}aturn tour images, raw ({PDS} volumes {COISS\_2xxx}) and calibrated products},
  howpublished = {NASA Planetary Data System, Cartography and Imaging Sciences Node and Ring-Moon Systems Node; retrieved through OPUS, \url{https://opus.pds-rings.seti.org}},
  note = {Accessed November 20, 2026}}

@article{Knowles2020,
  author = {B. Knowles and R. West and P. Helfenstein and A. Verbiscer and D. Wilson and C. Porco},
  title = {End-of-mission calibration of the {C}assini {I}maging {S}cience {S}ubsystem},
  journal = {Planet. Space Sci.}, volume = {185}, year = {2020}, pages = {104898},
  doi = {10.1016/j.pss.2020.104898}}

@article{Acton2018,
  author = {C. Acton and N. Bachman and B. Semenov and E. Wright},
  title = {A look towards the future in the handling of space science mission geometry},
  journal = {Planet. Space Sci.}, volume = {150}, year = {2018}, pages = {9--12},
  doi = {10.1016/j.pss.2017.02.013}}

@article{OzgurAlkhatibMezouarBartoli2026,
  author = {E. Ozgur and M. Alkhatib and Y. Mezouar and A. Bartoli},
  title = {Reconstructing a sphere and the camera focal length from a single view by fitting planes},
  journal = {Int. J. Comput. Vis.},
  volume = {134}, pages = {Art. 151}, year = {2026},
  doi = {10.1007/s11263-025-02697-9}
}

\appendix
\section{Proofs}\label{app:ellipsoids}

Write $(a_1,a_2,a_3)=(a,b,c)$, so $\Lam_i=a_i^{-2}$, and $D=\Lam^{-1}=\diag(a^2,b^2,c^2)$.

\paragraph{Notation for an image ellipse} We use $C$, $A_E$, $\alpha_1\le\alpha_2$, $\beta_E$ and $\gamma_E$ from \cref{sec:setting}; $\pi$, $q$, $r$, $J$, $h$ and $R_k$ from \cref{sec:choosef}; and $S$, $s_\pm$ and $\beta_\pm$ from \cref{sec:focal}, now with $S=A_E-\beta_E\beta_E\T/\gamma_E$ for every $\gamma_E\ne0$. Throughout, $t=f^2$. Write $\beta_E=(\beta_1,\beta_2)$ in an orthonormal eigenbasis of $A_E$ for $\alpha_1,\alpha_2$ (whereas $\beta_\pm$ refer to an eigenbasis of $S$). Expanding,
\[
\pi(\nu)=\gamma_E(\alpha_1-\nu)(\alpha_2-\nu)-\beta_1^2(\alpha_2-\nu)-\beta_2^2(\alpha_1-\nu),
\]
so $\pi(\alpha_1)=-(\alpha_2-\alpha_1)\beta_1^2\le0\le(\alpha_2-\alpha_1)\beta_2^2=\pi(\alpha_2)$, strictly when $p$ is off-axis; since $\pi$ has degree at most two, it then has exactly one root $r$ in $(\alpha_1,\alpha_2)$. For $\gamma_E\ne0$, $\pi(\nu)=\gamma_E\det(S-\nu I)$, so the roots of $\pi$ are the eigenvalues of $S$. If $\gamma_E>0$, then $\det S=\det C/\gamma_E<0$ and the other root of $\pi$ is negative. If $\gamma_E<0$ and $p$ is off-axis, the other root $r_2$ exceeds $\alpha_2$ (as $\pi(\alpha_2)>0$ and $\pi\to-\infty$), so $s_-=r$, $s_+=r_2$ and $\kappa_0^2=r_2/r$.

\subsection{The ratio test}

\begin{proof}[Proof of \cref{lem:cone}]
A direction $d$ is tangent if and only if $s\mapsto(O+sd)\T\Lam(O+sd)-1$ has a double root, that is, $(d\T\Lam O)^2=\varsigma\,d\T\Lam d$, which gives $M$. By the Sherman--Morrison formula, $(D-OO\T)^{-1}=\Lam+\Lam OO\T\Lam/(1-O\T\Lam O)=\Lam-\Lam OO\T\Lam/\varsigma$, so
\[
M=\varsigma\,(D-OO\T)^{-1}.
\]
The matrix $D-OO\T$ is the dual of the tangent cone: the image of the dual quadric $\diag(D,-1)$ under the camera $[\,I\mid-O\,]$ \cite[Sec.~8.3]{HartleyZisserman2004}. By the matrix determinant lemma,
\[
\det(D-OO\T+\lambda I)=P(\lambda):=\prod_i(a_i^2+\lambda)-\sum_iO_i^2\prod_{l\ne i}(a_l^2+\lambda),
\]
and $P(\lambda)/\prod_i(a_i^2+\lambda)$ is $1$ minus the left side of \eqref{eq:confocal}, so the roots $\lambda_1,\lambda_2,\lambda_3$ of $P$ are the confocal coordinates of $O$. Hence $D-OO\T$ has eigenvalues $-\lambda_1,-\lambda_2,-\lambda_3$ and $M$ has eigenvalues $-\varsigma/\lambda_i$; the $\lambda_i$ are nonzero because $P(0)=-a^2b^2c^2\varsigma<0$. Since $D-OO\T$ is a negative semidefinite rank-one update of $D$, Weyl's interlacing inequalities give $-\lambda_1\le c^2\le-\lambda_2\le b^2\le-\lambda_3\le a^2$, and $\lambda_1>0$ as $P(0)<0$. For $a>b>c$ and $O$ off the coordinate planes these inequalities are strict, since none of
\begin{gather*}
P(-a^2)=-O_1^2(a^2-b^2)(a^2-c^2),\qquad P(-b^2)=O_2^2(a^2-b^2)(b^2-c^2),\\
P(-c^2)=-O_3^2(a^2-c^2)(b^2-c^2)
\end{gather*}
vanishes.
\end{proof}

\begin{proof}[Proof of \cref{thm:ellipsoid}]
The matrix $Q_f$ defines the cone of rays through $E$, in camera coordinates. If $E$ is the ellipsoid's outline, the placed ellipsoid lies in $z<f$, so $O$ is outside it, and its semi-axes $sa\ge sb\ge sc$ have the same ratio $a/c$; we apply \cref{lem:cone} to it. The nappe of the tangent cone that contains the ellipsoid lies, apart from $O$, in $z<f$, and the other nappe in $z>f$; only the first meets the image plane. Moreover, the rays through the convex ellipsoid form a convex solid cone whose boundary rays are the tangent rays, so $E$ and the section of the tangent cone are the same real nondegenerate conic. Hence $UMU\T=\sigma_0Q_f$ for an orthogonal $U$ and some $\sigma_0\ne0$, and $\sigma_0>0$ since both matrices have two positive eigenvalues. So $\kappa(f)^2=\lambda_3/\lambda_2\le a^2/c^2$ by \cref{lem:cone}.

Conversely, let the cone $Q_f$, signed as in the statement, have eigenvalues $-\mu_0<0<\mu_-\le\mu_+$ with $\mu_+/\mu_-=k\in[1,a^2/c^2]$. We choose confocal coordinates $\lambda_i$ that reproduce the cone and then solve for $O$, an inverse eigenvalue problem for the rank-one update $D-OO\T$ \cite{Golub1973}; first let $a>b>c$.

\emph{Step 1 (confocal coordinates).} Put $\lambda_2=-\max(c^2,b^2/k)$ and $\lambda_3=-\max(kc^2,b^2)$. Then $-b^2\le\lambda_2\le-c^2$, $-a^2\le\lambda_3\le-b^2$ and $\lambda_3/\lambda_2=k$. Put $\lambda_1=|\lambda_3|\mu_-/\mu_0>0$.

\emph{Step 2 (the point $O$).} Set
\[
O_i^2=\frac{\prod_j(a_i^2+\lambda_j)}{\prod_{l\ne i}(a_i^2-a_l^2)},
\]
the classical solution of this inverse eigenvalue problem. For $a>b>c$ and $\lambda_1>0>\lambda_2\ge\lambda_3$, all three right sides are nonnegative if and only if $-a^2\le\lambda_3\le-b^2\le\lambda_2\le-c^2$. Indeed, given this chain, the factors $a_i^2+\lambda_j$ of the numerator are nonnegative for $j\le4-i$ and nonpositive for the other $i-1$ values of $j$, and the denominator has $i-1$ negative factors. Conversely, $i=2$ forces $\lambda_3\le-b^2\le\lambda_2$, and then $i=1$ forces $\lambda_3\ge-a^2$ and $i=3$ forces $\lambda_2\le-c^2$. By Step~1 the chain holds. By partial fractions, $\sum_iO_i^2/(a_i^2+\lambda)-1=-\prod_j(\lambda-\lambda_j)/\prod_i(a_i^2+\lambda)$, so $P(\lambda)=\prod_j(\lambda-\lambda_j)$ and $\lambda_1,\lambda_2,\lambda_3$ are the confocal coordinates of $O$. Since $P(0)=-\lambda_1\lambda_2\lambda_3<0$, we have $\varsigma>0$: $O$ lies outside the ellipsoid. (For $k=1$ and $k=a^2/c^2$ this gives the camera centers of \cref{prop:folds}.)

\emph{Step 3 (placing the camera).} By \cref{lem:cone}, $M$ has eigenvalues $-\varsigma/\lambda_1<0<\varsigma/|\lambda_3|\le\varsigma/|\lambda_2|$. By the choice of $\lambda_1$ and since $\lambda_3/\lambda_2=k$, these are $\sigma_0(-\mu_0,\mu_-,\mu_+)$ with $\sigma_0=\varsigma/(|\lambda_3|\mu_-)>0$. So $UMU\T=\sigma_0Q_f$ for an orthogonal $U$. Put the camera center at $O$ with camera coordinates $d=U(X-O)$. Since $d$ is proportional to $(u,v,f)$ at the image point $(u,v)$, a point $W$ in the coordinates of \cref{sec:setting} has $d=Z(W-(p,f))$ with $Z=\diag(1,1,-1)$; so $d_3>0$ means $z<f$, and the ellipsoid is placed by $X\mapsto(p,f)+ZU(X-O)$. Every orthogonal $U$ is realized by a rigid motion: if this map reverses orientation (that is, if $\det U=1$), replace $(U,O)$ by $(UZ,ZO)$; since $Z$ maps the ellipsoid to itself and $M(ZO)=ZM(O)Z$, the placed ellipsoid and $UMU\T$ are unchanged. A ray $d$ is then tangent to the ellipsoid if and only if $d\T Q_fd=0$, that is, since $Q_f=K_fCK_f$, if and only if its image point $\mathbf x=K_fd$ lies on $E$.

It remains to put the ellipsoid in front of the camera. On the plane $d_3=0$, $Q_f$ restricts to $f^2A_E\succ0$, so each nappe of the solid cone $\{d\T Q_fd\le0\}$ lies strictly on one side of that plane. The ellipsoid lies in the solid cone (the line $OX$ through any of its points $X$ meets it), is connected and misses $O$, so it lies in one nappe. If this is the nappe $d_3<0$, replace the ellipsoid by its point reflection through $O$; since an ellipsoid is centrally symmetric, this is a translate of the same ellipsoid (same orientation), and its tangent cone from $O$ is the same cone. Now for every $d\ne0$ in the nappe $d_3>0$ the ray $O+sU\T d$, $s>0$, meets the ellipsoid: the discriminant is nonnegative, and the two intersection parameters have positive product $\varsigma/(d\T U\Lam U\T d)$, hence the same sign, which is positive because the ellipsoid lies in this nappe. So the image of the ellipsoid is the region bounded by $E$; every point $\mathbf x$ of $E$ gives the generator $K_f^{-1}\mathbf x$, which has $d_3=1>0$, lies in this nappe and touches the ellipsoid, so the outline is exactly $E$.

\emph{Spheroids.} Only Step~2 needs $a>b>c$. For a spheroid the same construction, carried out in a plane through the symmetry axis, gives the camera center in closed form, and for a sphere any point at a suitable distance from the center works; the details are in the supplementary materials.

Part~(ii) follows from part~(i), which does not involve $b$.
\end{proof}

\begin{proof}[Proof of \cref{prop:folds}]
Let $a>b>c$. By \cref{lem:cone}, $\kappa^2=|\lambda_3|/|\lambda_2|$ with $|\lambda_2|\ge c^2$ and $|\lambda_3|\le a^2$. Equality in both means that $-c^2$ and $-a^2$ are roots of $P$, which by the values of $P(-c^2)$ and $P(-a^2)$ forces $O_3=O_1=0$: $O$ is on the middle axis. Conversely, for $O=(0,O_2,0)$ outside, $P=(a^2+\lambda)(c^2+\lambda)(b^2+\lambda-O_2^2)$, so $\lambda_2=-c^2$ and $\lambda_3=-a^2$; such $O$ are the points with $|O_2|>b$, so the bound is attained.

At fixed $f$, an outline with $\kappa=a/c$ is a boundary point of the set of conics with $\kappa\le a/c$. Adding a small positive multiple of $K_f^{-1}v_+v_+\T K_f^{-1}$ to $C$, with $v_+$ a unit eigenvector of $\mu_+$, adds the same multiple of $v_+v_+\T$ to $Q_f$ and raises $\mu_+$ alone, and the perturbed conic is still a real ellipse ($A_E$ stays positive definite and $\det C$ keeps its sign). So arbitrarily close to such an outline there are real ellipses that are not outlines at this $f$, and the outline is not an interior point of the image. The image map is smooth on the open set of placements (\cref{sec:test}), and by the submersion theorem it maps every neighborhood of a point where its differential is onto to a neighborhood of the image of that point, so every preimage of such an outline is a critical point of the image map.

For the other extreme, $|\lambda_3|/|\lambda_2|=1$ with $|\lambda_2|\le b^2\le|\lambda_3|$ forces $\lambda_2=\lambda_3=-b^2$, a double root of $P$. A root $-b^2$ forces $O_2=0$, and then $P(\lambda)=(b^2+\lambda)\bigl[(a^2+\lambda)(c^2+\lambda)-O_1^2(c^2+\lambda)-O_3^2(a^2+\lambda)\bigr]$; the bracket vanishes at $-b^2$ if and only if $O_1^2/(a^2-b^2)-O_3^2/(b^2-c^2)=1$. Conversely, on this hyperbola $-b^2$ is a double root, and since $\lambda_1>0$ it is $\lambda_2=\lambda_3$.

For a prolate spheroid ($b=c$), $P(\lambda)=(c^2+\lambda)\tilde P(\lambda)$ with $\tilde P(\lambda)=(a^2+\lambda)(c^2+\lambda)-O_1^2(c^2+\lambda)-(O_2^2+O_3^2)(a^2+\lambda)$, so $\lambda_2=-c^2$ always and $\kappa^2=|\lambda_3|/c^2$, where $\lambda_3\in[-a^2,-c^2]$ is a root of $\tilde P$. Since $\tilde P(-a^2)=O_1^2(a^2-c^2)$ and $\tilde P(-c^2)=-(O_2^2+O_3^2)(a^2-c^2)$, $\kappa=a/c$ exactly when $O_1=0$, and $\kappa=1$ exactly when $O_2=O_3=0$, outside the spheroid when $|O_1|>a$. The oblate case ($a=b$) is the same with the roles of $-a^2$ and $-c^2$ exchanged: $\lambda_3=-a^2$ always, $\kappa^2=a^2/|\lambda_2|$, $\kappa=a/c$ exactly when $O_3=0$, and $\kappa=1$ exactly when $O_1=O_2=0$.
\end{proof}

\subsection{Where the camera can be}

\begin{proof}[Proof of \cref{prop:confocal}]
Put $O=(x,y,z)$, $z=f>0$, in the frame of $E$.

\emph{The cone.} The planes tangent to $E$ are those with dual coordinates $(n,n_0)$ satisfying $a_E^2n_1^2+b_E^2n_2^2=n_0^2$, so $E$ is the degenerate dual quadric $\diag(D_E,-1)$, $D_E=\diag(a_E^2,b_E^2,0)$. As in the proof of \cref{lem:cone} (dual-quadric projection and the matrix determinant lemma, neither of which needs $D_E$ to be invertible), the cone of rays from $O$ through $E$ has dual $D_E-OO\T$, and
\[
G(\theta):=\det(D_E-OO\T+\theta I)=\theta(a_E^2+\theta)(b_E^2+\theta)\Bigl[1-\frac{x^2}{a_E^2+\theta}-\frac{y^2}{b_E^2+\theta}-\frac{z^2}{\theta}\Bigr]
\]
has the roots $\theta_i$. As $G(0)=-a_E^2b_E^2z^2<0$, the dual cone is nonsingular with eigenvalues $-\theta_i$, and the cone has eigenvalues proportional to $-1/\theta_i$; $Q_f$ differs from it by an orthogonal change of coordinates and a positive factor. Weyl interlacing gives the ranges of the $\theta_i$, as in \cref{lem:cone}. The positive pair is proportional to $1/|\theta_2|\ge1/|\theta_3|$, so $\kappa^2=\theta_3/\theta_2$.

\emph{The region.} Since $|\theta_3|\in[b_E^2,a_E^2]$, $|\theta_3|\le\rho^2|\theta_2|$ holds whenever $|\theta_2|\ge a_E^2/\rho^2$ and fails whenever $|\theta_2|<b_E^2/\rho^2$; this gives the two inclusions. On the plane $x=0$, $G(-a_E^2)=-x^2a_E^2(a_E^2-b_E^2)$ vanishes, so $\theta_3=-a_E^2$, and the region there is the inner set $\{|\theta_2|\ge a_E^2/\rho^2\}$. On the plane $y=0$, $G(-b_E^2)=y^2b_E^2(a_E^2-b_E^2)$ vanishes, and the boundary points with $\theta_3=-b_E^2$ have $\theta_2=-b_E^2/\rho^2$, so they lie on the outer hyperboloid. Along the boundary, $\theta_2$ takes every value in $[\max(-b_E^2,-a_E^2/\rho^2),-b_E^2/\rho^2]$, an interval of positive length for $\rho>1$, and $\theta_1>0$ is free. So no $\theta_i$ is constant on the boundary, and it is not a confocal quadric.

\emph{Circular cones and $f\to\infty$.} $\kappa=1$ forces the double root $\theta_2=\theta_3=-b_E^2$. As in the proof of \cref{prop:folds} with $(a,b,c)$ replaced by $(a_E,b_E,0)$, this means $y=0$ and $x^2/(a_E^2-b_E^2)-z^2/b_E^2=1$, the focal hyperbola of $E$. As $z\to\infty$, $G(\theta)/z^2\to-(a_E^2+\theta)(b_E^2+\theta)$, so the two negative roots tend to $-b_E^2$ and $-a_E^2$. On the plane $y=0$, $-b_E^2$ is a root, and the other negative root is the hyperbola label $\theta$; the formula for $\kappa^2$ after the proposition follows from the two orderings of $\theta$ and $-b_E^2$.
\end{proof}

\subsection{Unknown focal length}

\begin{proof}[Proof of \cref{prop:kmin}]
The first claim is \cref{thm:ellipsoid}(i) and the definition of an infimum. By \cref{thm:ellipsoid} with $a=c$, $\kappa(f)=1$ exactly when $E$ is a sphere outline at $f$. In the notation above,
\[
Q_f=\begin{pmatrix}f^2A_E&f\beta_E\\ f\beta_E\T&\gamma_E\end{pmatrix}.
\]
As $f\to\infty$, $Q_f/f^2\to\diag(A_E,0)$, whose positive eigenvalues are $\alpha_1,\alpha_2$, so $\kappa\to a_E/b_E$. As $f\to0$ there are three cases. If $\gamma_E>0$ ($p$ outside $E$), one eigenvalue tends to $\gamma_E$ and the other two are $f^2$ times the eigenvalues of $S$ plus $o(f^2)$; since $\det C=\gamma_E\det S<0$, $S$ has one positive eigenvalue, so the positive pair is $\gamma_E$ and $O(f^2)$, and $\kappa\to\infty$. If $\gamma_E=0$ ($p$ on $E$), then $\beta_E\ne0$ (because $\det C=-\beta_E\T\operatorname{adj}(A_E)\,\beta_E<0$), and $Q_f$ has eigenvalues $\pm f|\beta_E|+O(f^2)$ and a third eigenvalue $-f^2\det C/|\beta_E|^2+o(f^2)>0$, from $\det Q_f=f^4\det C$; so again $\kappa\to\infty$. If $\gamma_E<0$ ($p$ inside $E$), $S=A_E+\beta_E\beta_E\T/|\gamma_E|$ is positive definite, the positive pair is $f^2$ times its eigenvalues plus $o(f^2)$, and $\kappa\to\kappa_0=\sqrt{s_+/s_-}$.

Finally, since $\kappa$ is continuous on $(0,\infty)$ with limits at both ends, an unattained infimum equals one of the end limits; so $\kmin=1$ without attainment needs $\kappa\to1$ at an end. As $f\to\infty$ this happens exactly when $E$ is a circle, and then not centered at $p$, since otherwise $\kappa\equiv1$; conversely such a circle is never a sphere outline (\cref{lem:sphere}), so $\kappa>1$ for all $f$. As $f\to0$ it needs $p$ inside $E$ and $S=sI$ for some $s>0$, that is, $C=s\diag(1,1,0)-\ell\ell\T$ with $\ell=(\beta_E,\gamma_E)/\sqrt{|\gamma_E|}$. In pixel coordinates $(u,v)$ centered at $p$, the conic is then $s(u^2+v^2)=(\ell_1u+\ell_2v+\ell_3)^2$, the focus--directrix equation with focus $p$, so $p$ is a focus of $E$ (or the center, if $E$ is a circle centered at $p$). Conversely, if $p$ is a focus of a noncircular $E$, then $d_E=e_E$, so $\kappa>1$ for every $f>0$ by \cref{lem:sphere}; the focus--directrix equation gives $C=s\diag(1,1,0)-\ell\ell\T$ with $\ell_3\ne0$, so $\gamma_E=-\ell_3^2<0$ and $S=sI$, that is, $\kappa\to\kappa_0=1$ as $f\to0$.
\end{proof}

\paragraph{Sketch of the proof of \cref{thm:shape}(i)} The smaller root $u=\nu_-$ parametrizes $f$ monotonically (Steps 1--2); each crossing $\kappa^2=k$ is a root in $J=(\alpha_1,r)$ of a quartic $R_k$ (Step~3); sign changes of $R_k$ (\cref{tab:Rk}) force two or three roots outside $J$ (Step~4), leaving too few in $J$ for a disconnected sublevel set (Step~5); degenerate positions follow by perturbation (Step~6).

\begin{table}[htbp]
\centering\small
\caption{Signs of the quartic $R_k$ (Step 4) and the roots they force outside $J=(\alpha_1,r)$. Here $0<\alpha_1/k<\alpha_1<r<\alpha_2$ and $\kappa_\infty^2=\alpha_2/\alpha_1$.}\label{tab:Rk}
\setlength{\tabcolsep}{3pt}\footnotesize
\begin{tabular}{lccccccl}
 & $-\infty$ & $0$ & $\alpha_1/k$ & $\alpha_1$ & $\alpha_2$ & $+\infty$ & roots forced outside $J$\\\hline
$p$ outside $E$ & $-$ & $+$ & $-$ & $\operatorname{sgn}(k-\kappa_\infty^2)$ & $-$ & $-$ & $(-\infty,0)$, $(0,\alpha_1/k)$\\
$p$ inside $E$ & $+$ & $+$ & $-$ & $\operatorname{sgn}(k-\kappa_\infty^2)$ & $-$ & $+$ & $(0,\alpha_1/k)$, $(\alpha_2,\infty)$\\\hline
\multicolumn{8}{l}{and, in both cases, one more in $(\alpha_1/k,\alpha_1)$ when $k>\kappa_\infty^2$.}
\end{tabular}
\end{table}

\begin{proof}[Proof of \cref{thm:shape}(i)]
\emph{Step 1 (reduction).} With $N_t=\diag(1,1,t)$, $Q_f-\nu I=K_f(C-(\nu/t)N_t)K_f$, so the eigenvalues of $Q_f$ are $t$ times the roots of $\det(C-\nu N_t)=\pi(\nu)-t\nu q(\nu)$, and $\kappa(f)^2$ is the ratio of the two positive roots of this cubic. A root $\nu$ with $\nu q(\nu)\ne0$ satisfies $h(\nu)=t$; we call $t$ its \emph{height}.

\emph{Step 2 (the smaller root parametrizes $f$).} Assume first that $p$ is off-axis and not on $E$; the remaining cases are treated in Step~6. Cauchy interlacing of $Q_f$ with its block $tA_E$ gives $-\mu_0\le t\alpha_1\le\mu_-\le t\alpha_2\le\mu_+$; dividing by $t$, the two positive roots satisfy $\alpha_1<\nu_-<\alpha_2<\nu_+$, strictly, since $\alpha_i$ is a root of $\pi(\nu)-t\nu q(\nu)$ only if $\pi(\alpha_i)=0$, that is, $\beta_i=0$. The simple root $\nu_-(t)$ is continuous in $t$ and injective, since $h(\nu_-(t))=t$, hence strictly monotone. As $t\to\infty$, $Q_f/t$ tends to $\diag(A_E,0)$, so $\nu_-\to\alpha_1$; as $t\to0$, the equation reduces to $\pi(\nu)=0$, so $\nu_-\to r$. Hence $u=\nu_-$ is a decreasing bijection from $(0,\infty)$ onto $J=(\alpha_1,r)$, with inverse $h$.

\emph{Step 3 (the level equation).} For each $t$ the cubic $\det(C-\nu N_t)$ has the two positive roots $\nu_-,\nu_+$ and one negative root. So for $k>1$, $\kappa^2=k$ only if $ku$ is the other positive root at the same height, that is, only if $h(ku)=h(u)$. Since
\[
h(ku)-h(u)=\frac{(k-1)\,R_k(u)}{k\,u\,q(u)\,q(ku)},\qquad R_k(u)=\frac{k\,\pi(u)\,q(ku)-\pi(ku)\,q(u)}{1-k},
\]
this means $R_k(u)=0$: clearing denominators is legitimate, since $u\,q(u)\ne0$ on $J$ and $q(ku)\ne0$ for $ku=\nu_+>\alpha_2$. So for $k>1$, every $u\in J$ with $\kappa^2(u)=k$ is a root of the quartic $R_k$, and distinct such $u$ are distinct roots.

\emph{Step 4 (roots off $J$).} The leading coefficient of $R_k$ is $-\gamma_Ek^2$, and $R_k(0)=-\alpha_1\alpha_2\det C>0$ (so $R_k\not\equiv0$), $R_k(\alpha_1/k)=\pi(\alpha_1)q(\alpha_1/k)/(k-1)<0$, $R_k(\alpha_1)=-k\alpha_1\pi(\alpha_1)(k\alpha_1-\alpha_2)$, of the sign of $k-\kappa_\infty^2$, and $R_k(\alpha_2)=k\pi(\alpha_2)q(k\alpha_2)/(1-k)<0$, since $q(\alpha_1/k)>0$ and $q(k\alpha_2)>0$.
Since $R_k\to-\infty$ at both ends if $p$ lies outside $E$ and $R_k\to+\infty$ if $p$ lies inside, these signs force the roots listed in \cref{tab:Rk}: each is a root of odd multiplicity in an interval at whose ends $R_k$ has opposite signs, the intervals are pairwise disjoint, and none meets $J$. A quartic has four roots counted with multiplicity, so $J$ carries at most two roots of $R_k$, counted with multiplicity, when $k<\kappa_\infty^2$, and at most one when $k>\kappa_\infty^2$.

\emph{Step 5 (sublevel sets).} For $k>1$ let $g_k(u)=\kappa^2(u)-k$ on $J$; by Step~3, distinct zeros of $g_k$ are distinct roots of $R_k$. As $u\to\alpha_1$ ($f\to\infty$), $\kappa^2\to\kappa_\infty^2$ (\cref{prop:kmin}). Suppose that for some $k\ge1$ the set $\{\kappa^2\le k\}$ is disconnected: there are $u_1<u_2<u_3$ in $J$ with $\kappa^2(u_1),\kappa^2(u_3)\le k<\kappa^2(u_2)$. Choose $k'\in(k,\kappa^2(u_2))$ with $k'\ne\kappa_\infty^2$. By the intermediate value theorem, $g_{k'}$ has a zero in $(u_1,u_2)$ and one in $(u_2,u_3)$. If $k'<\kappa_\infty^2$, it has a third zero in $(\alpha_1,u_1)$, since $g_{k'}\to\kappa_\infty^2-k'>0$ as $u\to\alpha_1$. So $R_{k'}$ has at least three distinct roots in $J$ if $k'<\kappa_\infty^2$ and at least two if $k'>\kappa_\infty^2$, contradicting Step~4. Hence $\{\kappa^2\le k\}$ is an interval of $J$ for every $k\ge1$, and, as $u\mapsto f$ is monotone, $I(\sqrt k)$ is an interval.

\emph{Step 6 (circles and degenerate positions of $p$).} If $E$ is a circle or $p$ lies on an axis line of $E$, the parametrization of Step~2 degenerates; if $p$ lies on $E$, $R_k$ drops to a cubic and \cref{tab:Rk} changes. In all these cases we argue by perturbation. Suppose $I(\sqrt k)$ were not an interval: there are $f_1<f_2<f_3$ with $\kappa^2(f_1),\kappa^2(f_3)\le k<\kappa^2(f_2)$. Choose $k'\in(k,\kappa^2(f_2))$, so that $\kappa^2(f_1),\kappa^2(f_3)<k'<\kappa^2(f_2)$. For fixed $f$, $Q_f$ and hence $\kappa^2$ depend continuously on $C$, written in coordinates centered at $p$ (so moving $p$ changes $\beta_E$ and $\gamma_E$), as long as $A_E\succ0$ and $\det C<0$. So these strict inequalities persist for every nearby pair $(E',p')$ of an ellipse and a principal point, in particular for a noncircular $E'$ and a point $p'$ off $E'$ that is off-axis for $E'$. For that pair, Steps 1--5 show that $I(\sqrt{k'})$ is an interval, a contradiction.
\end{proof}

\begin{proof}[Proof of \cref{thm:shape}(ii)]
\emph{Generic $p$.} Suppose first that $p$ is off-axis and not on $E$, and use the notation of the proof of \cref{thm:shape}(i). Since $\nu_-$ is a simple root, $h'(u)\,d\nu_-/dt=1$. So $h'\ne0$ on $J$, $u=\nu_-$ parametrizes $f$ diffeomorphically, and critical points in $f$ and in $u$ coincide. Off the axis lines, $O$ is never on the focal hyperbola, so $\kappa^2>1$ is a smooth function of $u$. Write $R(k,u)=R_k(u)$, a polynomial in $(k,u)$ of degree exactly $4$ in $u$, since its leading coefficient $-\gamma_Ek^2$ is nonzero for $p\notin E$. By Step~3, $R(\kappa^2(u),u)=0$ on $J$. Differentiating, $\partial_kR\cdot(\kappa^2)'+\partial_uR=0$, so at a critical point $u_0$, with $k_0=\kappa^2(u_0)$, $u_0$ is a root of $R_{k_0}$ of multiplicity at least $2$.

Besides the signs in \cref{tab:Rk}, we need $R_k(r)=\pi(kr)\,q(r)/(k-1)$, where $q(r)<0$. If $p$ is outside $E$, the other root of $\pi$ is negative, so $\pi(kr)>0$ and $R_k(r)<0$. If $p$ is inside $E$, $\pi$ is positive exactly between $r$ and $r_2$, so $R_k(r)$ has the sign of $k-\kappa_0^2$, where $\kappa_0^2=r_2/r$. Recall also that $R_k(\alpha_1)$ has the sign of $k-\kappa_\infty^2$, where $\kappa_\infty^2=\alpha_2/\alpha_1$ is the limit of $\kappa^2$ at the end $u=\alpha_1$.

Count the roots of $R_{k_0}$ with multiplicity. Each root below is forced in an interval at whose ends $R_{k_0}$ has opposite signs or vanishes, and these intervals are pairwise disjoint and do not contain $u_0$, so the roots are distinct. Besides the double root $u_0$, \cref{tab:Rk} forces two roots outside $J$: in $(-\infty,0)$ and $(0,\alpha_1/k_0)$ for $p$ outside $E$, and in $(0,\alpha_1/k_0)$ and $(\alpha_2,\infty)$ for $p$ inside. One more root is forced, in a set disjoint from these and from $J$, in each of the following cases: if $k_0\ge\kappa_\infty^2$, a root in $(\alpha_1/k_0,\alpha_1]$, since $R_{k_0}(\alpha_1/k_0)<0\le R_{k_0}(\alpha_1)$; if $p$ is inside $E$ and $k_0\ge\kappa_0^2$, a root in $[r,\alpha_2)$, since $R_{k_0}(r)\ge0>R_{k_0}(\alpha_2)$. Otherwise both end limits of $\kappa^2$ exceed $k_0$; if $u_0$ were not a global minimizer, there would be $\bar u$ with $\kappa^2(\bar u)<k_0$, and the level $k_0$ would be crossed again between $\bar u$ and the end of $J$ away from $u_0$, giving a further root in $J$. In each case $R_{k_0}$ would have at least $5$ roots, more than its degree $4$. So every critical point is a global minimizer, at a level below $\kappa_\infty$ and, for $p$ inside $E$, below $\kappa_0$.

Finally, suppose that $\kappa$ had two critical points. Both would be global minimizers. Since the set of minimizers of the quasiconvex $\kappa$ is an interval, $\kappa$ would be constant on a nondegenerate interval, and hence, being real-analytic (its positive eigenvalues are simple), on all of $(0,\infty)$. Its end limits would then equal its critical value, which we have excluded. So $\kappa$ has at most one critical point, and away from it $\kappa'\ne0$, which gives the monotonicity.

\emph{$p$ on $E$, off-axis.} Then $\gamma_E=0$, and $\pi(\nu)=(\beta_1^2+\beta_2^2)\nu-(\beta_1^2\alpha_2+\beta_2^2\alpha_1)$ is linear and increasing, with its root $r$ in $(\alpha_1,\alpha_2)$. Steps 1--3 of the proof of \cref{thm:shape}(i) still apply. In Step~2, two roots now escape like $\pm|\beta_E|/\sqrt t$, so still $\nu_-\to r$, and $\kappa\to\infty$ as $u\to r$ by \cref{prop:kmin}. Now $R_k$ is a cubic with leading coefficient $-(\beta_1^2+\beta_2^2)k(k+1)<0$. Its signs at $0$, $\alpha_1/k$, $\alpha_1$ and $\alpha_2$ are as before, and $R_k(r)=\pi(kr)q(r)/(k-1)<0$, since $\pi$ is linear and increasing with root $r$ and $kr>r$, so $\pi(kr)>0$, while $q(r)<0$ for $\alpha_1<r<\alpha_2$. Now only the root in $(0,\alpha_1/k_0)$ is forced in general, and the same count gives at least $4$ roots of the cubic in each case; the rest of the argument is unchanged.

\emph{$p$ on an axis line of $E$, or $E$ a circle.} Choose the eigenbasis of $A_E$ so that the center of $E$ lies on the line through $p$ in direction $e_1$. For a circle every orthonormal basis is an eigenbasis; for $p$ on the short axis line, exchange the indices $1$ and $2$ below. Then $\beta_E=(\beta_1,0)$ with $\beta_1\ne0$, since $p$ is not the center, and $Q_f$ splits: $e_2$ is an eigenvector with eigenvalue $f^2\alpha_2$, and the other two eigenvalues are those of $\left(\begin{smallmatrix}f^2\alpha_1&f\beta_1\\ f\beta_1&\gamma_E\end{smallmatrix}\right)$, whose determinant is $f^2\Delta$ with $\Delta=\alpha_1\gamma_E-\beta_1^2=\det C/\alpha_2<0$. With $\xi=1/f^2$, the positive one is $f^2\nu(\xi)$, where $\nu(\xi)$ is the simple positive root of $\Phi(\nu,\xi)=\nu^2-(\alpha_1+\gamma_E\xi)\nu+\xi\Delta$. If $\nu'(\xi)=0$, then $\partial\Phi/\partial\xi=\Delta-\gamma_E\nu=0$. This is impossible if $\gamma_E=0$; otherwise $\nu=\Delta/\gamma_E$ and $\Phi(\Delta/\gamma_E,\xi)=(\Delta/\gamma_E)(\Delta/\gamma_E-\alpha_1)=-\Delta\beta_1^2/\gamma_E^2\ne0$, a contradiction. So $\nu$ is strictly monotone in $f$, and $\kappa^2=\max(\nu/\alpha_2,\alpha_2/\nu)$. Hence $\kappa$ has no critical point, and it has a corner exactly where $\nu=\alpha_2$, that is, $\kappa=1$. This happens only for a sphere outline (\cref{lem:sphere}), and a circle not centered at $p$ is never one: for a circle $\alpha_1=\alpha_2$, and $\Phi(\alpha_1,\xi)=-\xi\beta_1^2\ne0$, so $\nu\ne\alpha_2$. (Solving $\Phi(\nu,\xi)=0$ at $\nu\in\{\rho^2\alpha_2,\alpha_2/\rho^2\}$, where $\kappa=\rho$, gives $f^2=1/\xi=(\gamma_E\nu-\Delta)/(\nu(\nu-\alpha_1))$, the axis formula of \cref{alg:test}.)
\end{proof}

\begin{proof}[Proof of \cref{thm:shape}(iii)]
By \cref{thm:shape}(ii), and since $E$ is not a sphere outline from an off-axis $p$ (\cref{lem:sphere}), $\kappa$ is either strictly monotone or strictly decreasing and then strictly increasing. So $\kmin$ is attained exactly when $\kappa<\kappa_\infty$ for all large $f$ and $\kappa<\lim_{f\to0}\kappa$ for all small $f$. For large $f$ this means $|x|>|y|$, by \cref{prop:wedge}. For small $f$ it is automatic for $p$ outside or on $E$, where $\kappa\to\infty$ (\cref{prop:kmin}). For $p$ inside $E$, the expansion below gives $\kappa<\kappa_0$ for small $f$ if $\beta_+^2>\beta_-^2$ and $\kappa>\kappa_0$ for small $f$ if $\beta_+^2<\beta_-^2$; the case $\beta_+^2=\beta_-^2$, in which $\kmin$ is not attained, is treated at the end. If $\kmin$ is not attained, $\kappa$ has no critical point, so by \cref{thm:shape}(ii) it is strictly monotone and $\kmin$ is the smaller of its end limits.

\emph{Expansion at $f=0$.} For $p$ inside $E$,
\[
Q_f=\begin{pmatrix}tA_E&\sqrt t\,\beta_E\\ \sqrt t\,\beta_E\T&\gamma_E\end{pmatrix}.
\]
The two positive eigenvalues are $t\nu_\pm$ with $\nu_\pm\to s_\pm$ as $t\to0$ (proof of \cref{prop:kmin}), and $s_+>s_-$, since $S=sI$ only when $p$ is a focus, which lies on an axis line. They make the Schur complement with respect to the last entry singular: $\nu_\pm$ is an eigenvalue of $A_E-\beta_E\beta_E\T/(\gamma_E-t\nu_\pm)=S-\chi_0(\nu_\pm)\,\beta_E\beta_E\T$, where $\chi_0(\nu)=1/(\gamma_E-t\nu)-1/\gamma_E=t\nu/\gamma_E^2+O(t^2)$; by the implicit function theorem applied to $\det(S-\chi_0(\nu)\beta_E\beta_E\T-\nu I)=0$ at the simple roots $(t,\nu)=(0,s_\pm)$, $\nu_\pm$ are smooth in $t$. First-order perturbation of the simple eigenvalues $s_\pm$ of $S$ gives $\nu_\pm=s_\pm-\chi_0(\nu_\pm)\beta_\pm^2+O(t^2)$, and $\chi_0(\nu_\pm)=ts_\pm/\gamma_E^2+O(t^2)$, so $\nu_\pm=s_\pm(1-t\beta_\pm^2/\gamma_E^2)+O(t^2)$, so
\[
\kappa^2=\frac{\nu_+}{\nu_-}=\kappa_0^2\Bigl(1-\frac{\beta_+^2-\beta_-^2}{\gamma_E^2}\,t\Bigr)+O(t^2),\qquad \kappa_0^2=s_+/s_-,
\]
the expansion stated after \cref{thm:shape}.

\emph{The case $\beta_+^2=\beta_-^2$.} The sign of $\beta_+^2-\beta_-^2$ does not depend on the scale of $C$; normalize $C$ as in the proof of \cref{prop:wedge}, so $A_E=\diag(1/A,1/B)$, $\beta_E=(x/A,y/B)$ and $\gamma_E=x^2/A+y^2/B-1$, with $A=a_E^2$, $B=b_E^2$. Since $S$ is a symmetric $2\times2$ matrix, $\beta_+^2-\beta_-^2=(2\beta_E\T S\beta_E-\operatorname{tr}S\,|\beta_E|^2)/(s_+-s_-)$, and a direct computation gives
\[
\begin{gathered}
\beta_+^2-\beta_-^2=\frac{N(x^2,y^2)}{A^3B^3(1-x^2/A-y^2/B)(s_+-s_-)},\\
N(X,Y)=(A-B)(A^2Y-B^2X)+(AY+BX)^2-(A-B)^2XY.
\end{gathered}
\]
The conic $N(X,Y)=0$ is nondegenerate (the determinant of its $3\times3$ matrix is $-A^3B^3(A-B)^2\ne0$), so the gradient of $N$ does not vanish on it; as $(x,y)\mapsto(x^2,y^2)$ is a local diffeomorphism off the axes, every off-axis $p$ inside $E$ with $\beta_+^2=\beta_-^2$ is a limit of off-axis points $p'$ inside $E$ with $\beta_+'^2<\beta_-'^2$. For such $p'$, $\kappa>\kappa_0$ for small $f$, so $\kappa$ is not decreasing near $0$. As $p'$ is off-axis, $E$ is not a sphere outline from it (\cref{lem:sphere}); by \cref{thm:shape}(ii), $\kappa$ therefore has no critical point (a critical point is preceded by a strict decrease) and increases strictly, and $\kappa(f;p')>\kappa_0(p')$ for every $f$. Since $\kappa(f;\cdot)$ and $\kappa_0$ are continuous inside $E$, $\kappa(f;p)\ge\kappa_0(p)$ for every $f$. Off the axes $\kappa>1$ is smooth, so an attained minimum is a critical point and lies below $\kappa_0$ (\cref{thm:shape}(ii)); hence $\kmin$ is not attained.
\end{proof}

\begin{proof}[Proof of \cref{prop:wedge}]
Normalize $C$ as in \cref{thm:kminform} and write it in the eigenbasis of $A_E$, so $\alpha_1=1/a_E^2<\alpha_2=1/b_E^2$ and, with $(x,y)$ the position of $p$ in the frame of $E$, $\beta_E=(\beta_1,\beta_2)=(\alpha_1x,\alpha_2y)$ and $\gamma_E=\alpha_1x^2+\alpha_2y^2-1$. With $\eta=1/f$,
\[
Q_f/f^2=\diag(\alpha_1,\alpha_2,0)+\eta\begin{pmatrix}0&\beta_E\\ \beta_E\T&0\end{pmatrix}+\eta^2\diag(0,0,\gamma_E).
\]
The eigenvalues $\alpha_1\ne\alpha_2$ are simple. Conjugating by $\diag(1,1,-1)$ changes the sign of $\eta$, so the eigenvalues are even functions of $\eta$. Since $\det(Q_f/f^2)=\eta^2\det C<0$, the eigenvalue near $0$ is negative, so for small $\eta$ the two positive eigenvalues $\mu_-/f^2<\mu_+/f^2$ of $Q_f/f^2$ are the perturbations of $\alpha_1$ and $\alpha_2$, and $\kappa^2=\mu_+/\mu_-$. Second-order perturbation theory, in which the first-order terms and the coupling between the first two eigenvectors vanish, gives
\begin{gather*}
\mu_-/f^2=\alpha_1+\eta^2\beta_1^2/\alpha_1+O(\eta^4)=\alpha_1(1+x^2\eta^2)+O(\eta^4),\\
\mu_+/f^2=\alpha_2(1+y^2\eta^2)+O(\eta^4).
\end{gather*}
Dividing gives the expansion.

In case~(i), $\kappa<\kappa_\infty$ for large $f$, so $\kmin<\kappa_\infty$, and for $\rho\in(\kmin,\kappa_\infty)$ the set $I(\rho)$ is nonempty and bounded above, since $\kappa\to\kappa_\infty>\rho$ as $f\to\infty$ (\cref{prop:kmin}).

In case~(ii), let $|x|\le|y|$; then $y\ne0$. Write $A=a_E^2$, $B=b_E^2$. If $x=0$, the proof of \cref{prop:confocal} gives $\theta_3=-A$ and $\theta_2\in(-B,0)$, so $\kappa^2=\theta_3/\theta_2>A/B=\kappa_\infty^2$ for every $f$. Now let $p$ be off-axis, and suppose $\kappa(f)=\kappa_\infty$ for some $f>0$. By \cref{prop:confocal}, $\theta_2=-sB$ and $\theta_3=-sA$ for some $s$, and $s\in(B/A,1)$ because $-B<\theta_2<0$ and $-A<\theta_3<-B$ off the axes. Both are roots of $x^2/(A+\theta)+y^2/(B+\theta)+f^2/\theta=1$, so both values of $\theta\bigl(1-x^2/(A+\theta)-y^2/(B+\theta)\bigr)$ equal $f^2$. With $m=As-B>0$ and $n=A-Bs>0$, the value at $\theta_2$ minus that at $\theta_3$ is
\[
s(A-B)\,\frac{(1-s)mn-x^2m+y^2n}{(1-s)mn},
\]
whose numerator is positive, since $x^2m-y^2n\le y^2(m-n)=-(A+B)(1-s)y^2<0$. So $\kappa\ne\kappa_\infty$ on the set of $(p,f)$ with $p$ off-axis, $|x|\le|y|$ and $f>0$. Each of its four connected components (one per quadrant) contains points with $p$ outside $E$, where $\kappa\to\infty$ as $f\to0$ (\cref{prop:kmin}), and $\kappa$ is continuous in $(p,f)$, since $Q_f$ is congruent to $C$ and so always has two positive eigenvalues; hence $\kappa>\kappa_\infty$ on the whole set. This proves case~(ii); the claims about $I(\rho)$ follow.
\end{proof}

\subsection{The minimizing focal length and \texorpdfstring{$\kmin$}{kmin} in closed form}

\begin{proof}[Proof of \cref{prop:fstar}]
Write $K=K_f=\diag(\sqrt t,\sqrt t,1)$, so $Q_f=KCK$ and $dK/dt=(K-e_3e_3\T)/(2t)$. Since $K^{-1}e_3=e_3$, $e_3e_3\T CK=e_3e_3\T Q_f$ and $KCe_3e_3\T=Q_fe_3e_3\T$, so
\[
\frac{dQ_f}{dt}=\frac{1}{2t}\bigl(2Q_f-e_3e_3\T Q_f-Q_fe_3e_3\T\bigr).
\]
For a simple eigenvalue $\mu$ with unit eigenvector $v$, $d\mu/dt=v\T(dQ_f/dt)v=\mu(1-(v\cdot e_3)^2)/t$. Hence $d\ln\mu_\pm/dt=(1-\cos^2\varphi_\pm)/t$, and subtracting gives $d\ln(\mu_+/\mu_-)/dt=(\cos^2\varphi_--\cos^2\varphi_+)/t$, that is, $d\ln\kappa/d\ln f=\cos^2\varphi_--\cos^2\varphi_+$. An attained $\kmin>1$ is an interior minimum of $\kappa$, which is differentiable there since $\mu_-<\mu_+$, hence a stationary point.

For the confocal form, work in the frame of $E$, with $O=(x,y,z)$, $z=f$, $A=a_E^2$ and $B=b_E^2$. Since $p$ is off-axis, the $\theta_i$ are simple roots of the confocal equation of \cref{sec:where}. By the proof of \cref{prop:confocal}, the dual cone $D_E-OO\T$ has eigenvalue $-\theta_i$ with eigenvector $v_i\propto(D_E+\theta_iI)^{-1}O=(x/(A+\theta_i),\,y/(B+\theta_i),\,z/\theta_i)$. The cone, a multiple of its inverse, has the same eigenvectors, and $Q_f$ differs from it by a positive factor and an orthogonal map that preserves the vertical line, so the values of $\cos^2$ are unchanged. The vector $\theta_iv_i$ has squared length $g(\theta_i)+z^2$ and vertical component $z$, so its angle $\varphi_i$ with the vertical satisfies $\cos^2\varphi_i=z^2/(g(\theta_i)+z^2)$. Since $|\theta_3|\ge|\theta_2|$, the eigenvector of $\mu_-$ belongs to $\theta_3$, so $\varphi_-=\varphi_3$ and $\varphi_+=\varphi_2$, and $\cos\varphi_-=\cos\varphi_+$ exactly when $g(\theta_2)=g(\theta_3)$.
\end{proof}

\begin{proof}[Proof of \cref{cor:lip}]
By \cref{prop:fstar}, $d\ln\kappa/d\ln f=\cos^2\varphi_--\cos^2\varphi_+\in[-1,1]$ wherever $\mu_-<\mu_+$; where $\mu_-=\mu_+$, $\kappa=1$ is its minimum value. The eigenvalues of $Q_f$ are locally Lipschitz in $f$ and $\mu_->0$, so $\ln\kappa$ is locally Lipschitz in $\ln f$, hence absolutely continuous, with $|d\ln\kappa/d\ln f|\le1$ almost everywhere; integrate in $\ln f$. For the inclusion, $\kappa(f)\le\kmin\max(f/f^*,f^*/f)\le\rho$ on the stated interval.
\end{proof}

\begin{proof}[Proof of \cref{prop:endpoints}]
At an endpoint $f\in(0,\infty)$ of $I(\rho)$, $\kappa(f)=\rho$ by continuity, so $u=\nu_-(f^2)\in J$ satisfies $\kappa^2(u)=\rho^2$ and is therefore a root of $R_{\rho^2}$ (Step~3 of the proof of \cref{thm:shape}(i)), and $f^2=h(u)$ (Step~2). For $p$ on $E$, Steps 1--3 hold as well (proof of \cref{thm:shape}(ii)). Conversely, if $u\in J$ and $R_k(u)=0$ with $k=\rho^2$, then $q(ku)\ne0$ (the roots of $q$ are $\alpha_1$ and $\alpha_2$, and $ku>u>\alpha_1$; at $ku=\alpha_2$, $R_k(u)=\pi(\alpha_2)q(u)/(k-1)\ne0$), so $h(ku)=h(u)$, $ku=\nu_+$ and $\kappa^2=k$ at $f=\sqrt{h(u)}$. If $f$ were interior to $I(\rho)$, then $\kappa\le\rho=\kappa(f)$ near $f$, so $f$ would be a local maximum of the smooth function $\kappa$ ($p$ is off-axis), hence a critical point; by \cref{thm:shape}(ii) a critical point is a strict minimum, a contradiction. So $f$ is an endpoint.
\end{proof}

\begin{proof}[Proof of \cref{thm:kminform}]
We use the notation of the proofs of \cref{thm:shape}(i) and (ii). Let $k^*=\kmin^2$ be attained at $u^*\in J$; $k^*>1$, since $E$ is not a sphere outline from an off-axis $p$ (\cref{lem:sphere}). By the proof of \cref{thm:shape}(ii), $u^*$ is a root of $R_{k^*}$ of multiplicity at least $2$, so $\mathcal D(k^*)=0$, and $f^{*2}=h(u^*)$ by Step~2. Since $k^*<\kappa_\infty^2$ and, for $p$ inside $E$, $k^*<\kappa_0^2$ (\cref{thm:shape}(ii)), the signs of \cref{tab:Rk} and of $R_{k^*}(r)$ give two roots of $R_{k^*}$ outside $J$ (one when $p$ lies on $E$, where $R_{k^*}$ is a cubic), so $u^*$ is exactly a double root.

Now let $k>k^*$. By \cref{thm:shape}(ii), $\kappa^2$ decreases strictly on $J$ from its limit at $u=r$ ($+\infty$ if $p$ is outside $E$, $\kappa_0^2$ if inside) to $k^*$ at $u^*$, then increases strictly to $\kappa_\infty^2$ at $u=\alpha_1$. So each end of $J$ contributes at least one root of $R_k$ (Step~3 and the signs of \cref{tab:Rk} and of $R_k(r)$): the end $u=\alpha_1$ a crossing in $(\alpha_1,u^*)$ if $k<\kappa_\infty^2$ and otherwise a root in $(\alpha_1/k,\alpha_1]$; the end $u=r$ a crossing in $(u^*,r)$ if its limit exceeds $k$ and otherwise a root in $[r,\alpha_2)$. The other two roots are in $(-\infty,0)$ and $(0,\alpha_1/k)$ if $p$ is outside $E$, and in $(0,\alpha_1/k)$ and $(\alpha_2,\infty)$ if $p$ is inside. These four roots lie in pairwise disjoint sets, so all roots of $R_k$ are simple and $\mathcal D(k)\ne0$. Hence $k^*$ is the largest root of $\mathcal D$ in $(1,\infty)$.

The substitution $u\mapsto ku$, $k\mapsto1/k$ maps $R_k$ to itself: $R_{1/k}(ku)=R_k(u)$, by the formula of Step~3. If $R_k$ has leading coefficient $c_k$ and roots $u_i$, then $R_{1/k}$ has leading coefficient $k^{-4}c_k$ and roots $ku_i$, so $\mathcal D(1/k)=(k^{-4}c_k)^6\prod_{i<j}(ku_i-ku_j)^2=k^{-12}\mathcal D(k)$. The symbolic computation described below shows that, in the normalization of the theorem, $\mathcal D(k)=(a_Eb_E)^{-22}k^2P_8(k)$ with $P_8$ of degree $8$; the reciprocity just proved then gives $P_8(1/k)=k^{-8}P_8(k)$, so $P_8$ is palindromic and $P_8(k)=k^4\Omega_{E,p}(k+1/k)$ for a quartic $\Omega_{E,p}$. The factorization and the coefficients $\omega_i$ (\cref{app:quartic}) are computed by the supplementary scripts \texttt{derive\_kmin\_quartic.py} and \texttt{verify\_discriminant\_identity.py}, which checks that the discriminant of $A^2B^2R_k(u)$ equals $ABk^2P_8(k)$ as an identity of integer polynomials in $A,B,x,y,k$; \texttt{verify\_discriminant\_numeric.py} checks it independently, without symbolic factorization, by computing the discriminant from $60$-digit roots at random rational points. Only this identity is computer-assisted. Finally, every real root $w>2$ of $\Omega_{E,p}$ is $k+1/k$ for a root $k>1$ of $P_8$, hence of $\mathcal D$, while real roots $w\le2$ come from $|k|=1$ or $k<0$. Since $w=k+1/k$ is increasing on $(1,\infty)$ and $k^*>1$, $w^*=k^*+1/k^*$ is the largest real root of $\Omega_{E,p}$, and $k^*$ is recovered from $w^*$ by solving a quadratic.

If $p$ lies on $E$, then $\gamma_E=0$ and $R_k$ is a cubic with nonzero leading coefficient (proof of \cref{thm:shape}(ii)). The argument above applies with three roots instead of four: for $k>k^*$ there is a root in $(0,\alpha_1/k)$, one in $(\alpha_1/k,\alpha_1]$ if $k\ge\kappa_\infty^2$, and the crossings in $J$, one on the branch toward $u=r$, where $\kappa\to\infty$, and one on the other branch if $k<\kappa_\infty^2$. The formal discriminant of a quartic with vanishing leading coefficient equals the square of the next coefficient times the discriminant of the cubic; here that coefficient is $-(\beta_1^2+\beta_2^2)k(k+1)$, so $\mathcal D=(\beta_1^2+\beta_2^2)^2k^2(k+1)^2\operatorname{disc}(R_k)$. The extra factor has no root in $(1,\infty)$, so $\mathcal D$ and $\operatorname{disc}(R_k)$ have the same roots there. Hence the largest root of $\Omega_{E,p}$ is still $k^*+1/k^*$; on $E$, $\Omega_{E,p}$ also has the root $w=-2$, from $(k+1)^2$.
\end{proof}

\section{Relation to the test of Gaudilli\`ere, Simon and Berger}\label{app:gsb}

For a triaxial ellipsoid and a calibrated camera, \cite[Thm.~1]{Gaudilliere2023} scans a scalar parameter and accepts when, for some value, three entries equal to the squared coordinates of the camera center are nonnegative. These camera centers are those of our Step~2 (proof of \cref{thm:ellipsoid}) with $(\lambda_1,\lambda_2,\lambda_3)=(\tau/\mu_0,-\tau/\mu_+,-\tau/\mu_-)$, $\tau>0$, their parameter being, up to sign, an increasing function of $\tau$. By Step~2 the entries are nonnegative exactly when $c^2\mu_+\le\tau\le b^2\mu_+$ and $b^2\mu_-\le\tau\le a^2\mu_-$, and such $\tau$ exists exactly when $c^2\mu_+\le a^2\mu_-$, which is our test. The supplementary materials give the details; \cref{sec:numerics} compares the two tests numerically.

\section{The quartic \texorpdfstring{$\Omega_{E,p}$}{Omega}}\label{app:quartic}

Let $A=a_E^2$, $B=b_E^2$, let $(x,y)$ be the principal point in the frame of $E$, $X=x^2$ and $Y=y^2$. Each coefficient $\omega_i$ of $\Omega_{E,p}(w)=\sum_{i=0}^4\omega_iw^i$ in \cref{thm:kminform} is a homogeneous polynomial of degree $10$ in $(A,B,X,Y)$ with integer coefficients, so rescaling the image multiplies $\Omega_{E,p}$ by a positive constant and does not change its roots; we list them in units with $b_E=1$ ($B=1$, $A=a_E^2/b_E^2$, $X=x^2/b_E^2$, $Y=y^2/b_E^2$), and the general coefficients are $B^{10}\omega_i(A/B,1,X/B,Y/B)$. The leading coefficient is $\omega_4=A^3B^3(A-B)^2|p-F_1|^2|p-F_2|^2$ (\cref{thm:kminform}). With $G=X/A+Y/B-1$ ($=\gamma_E$ in the normalization of \cref{thm:kminform}) and $H=(A(A-X)-B(B-Y))^2+4ABXY$,
\[
\Omega_{E,p}(-2)=-16AB\,G\,H^2,
\]
which vanishes for $p$ on $E$, and $\Omega_{E,p}(2)$ is $16XY(A-B)^2$ times a polynomial, so it vanishes for $p$ on an axis line. The coefficients $\omega_0,\dots,\omega_3$ have $105$, $101$, $87$ and $48$ terms; they are listed in the supplementary materials and, in machine-readable form, in the file \texttt{kmin\_quartic.txt} of the code repository, produced by \texttt{derive\_kmin\_quartic.py} and checked by \texttt{verify\_discriminant\_identity.py}.

\end{document}